%% file: main_arxiv.tex
\documentclass{article}

\usepackage[nonatbib,preprint]{neurips_2022}

\usepackage[utf8]{inputenc} 
\usepackage[T1]{fontenc}    
\usepackage{hyperref}       
\usepackage{url}            
\usepackage{booktabs}       
\usepackage{amsfonts}       
\usepackage{nicefrac}       
\usepackage{microtype}      
\usepackage{amssymb,amsmath,amsthm}
\usepackage{booktabs}
\usepackage{upgreek}
\usepackage{graphicx}
\usepackage{xcolor}
\usepackage{caption}
\usepackage{subcaption}
\usepackage{multirow}
\usepackage{enumitem}
\usepackage[ruled,vlined,linesnumbered]{algorithm2e}
\usepackage{xcolor}
\usepackage{multirow}
\usepackage{stackengine}
\usepackage{mathtools}
\usepackage[hang,flushmargin]{footmisc}
\usepackage[numbers]{natbib}

\newtheorem{definition}{Definition}

\DeclareMathOperator\artanh{artanh}
\DeclareMathOperator\arcosh{arcosh}

\definecolor{our_red}{RGB}{215, 103, 103}
\definecolor{our_blue}{RGB}{109, 177, 255}
\title{Towards Scalable Hyperbolic Neural Networks using Taylor Series Approximations}

\author{Nurendra Choudhary, Chandan K. Reddy\\
Department of Computer Science, Virginia Tech, Arlington, VA, USA\\
\texttt{nurendra@vt.edu, reddy@cs.vt.edu}}

\begin{document}

\maketitle

\input{abstract.tex}

\input{introduction.tex}

\input{related.tex}

\input{model.tex}

\input{experiments.tex}

\input{conclusion.tex}

\input{broader_impact.tex}

\bibliographystyle{plainnat}

\bibliography{neurips_2022}

\input{appendix.tex}

\end{document}

%% file: abstract.tex
\begin{abstract}
  Hyperbolic networks have shown prominent improvements over their Euclidean counterparts in several areas involving hierarchical datasets in various domains such as computer vision, graph analysis, and natural language processing. However, their adoption in practice remains restricted due to (i) non-scalability on accelerated deep learning hardware, (ii) vanishing gradients due to the closure of hyperbolic space, and (iii) information loss due to frequent mapping between local tangent space and fully hyperbolic space. To tackle these issues, we propose the approximation of hyperbolic operators using Taylor series expansions, which allows us to reformulate the computationally expensive tangent and cosine hyperbolic functions into their polynomial equivariants which are more efficient. This allows us to retain the benefits of preserving the hierarchical anatomy of the hyperbolic space, while maintaining the scalability over current accelerated deep learning infrastructure. The polynomial formulation also enables us to utilize the advancements in Euclidean networks such as gradient clipping and ReLU activation to avoid vanishing gradients and remove errors due to frequent switching between tangent space and hyperbolic space. Our empirical evaluation on standard benchmarks in the domain of graph analysis and computer vision shows that our polynomial formulation is as scalable as Euclidean architectures, both in terms of memory and time complexity, while providing results as effective as hyperbolic models. Moreover, our formulation also shows a considerable improvement over its baselines due to our solution to vanishing gradients and information loss.
\end{abstract}

%% file: introduction.tex
\section{Introduction}
\label{sec:intro}
\begin{figure}[htbp]
\vspace{-1em}
    \centering
    \begin{subfigure}{.25\textwidth}
    \centering
    \includegraphics[width=\linewidth]{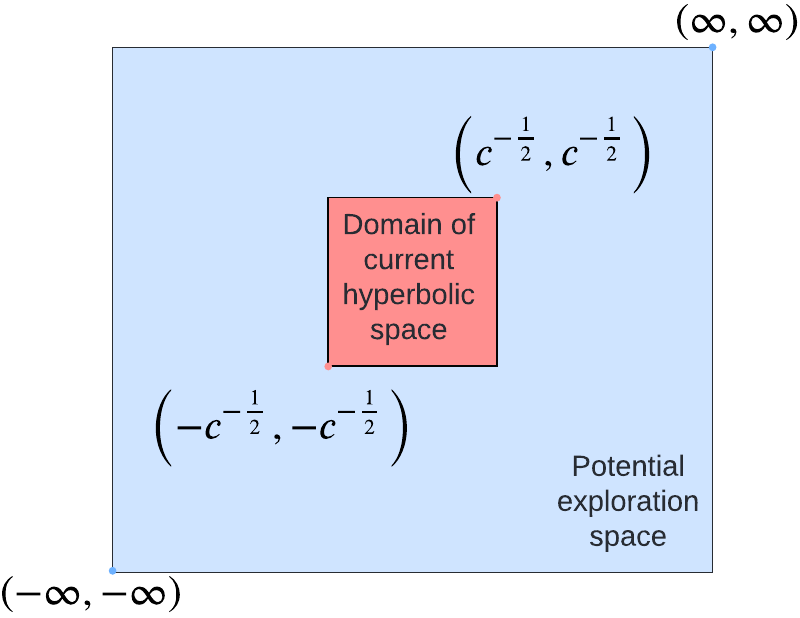}
    \vspace{-1em}
    \caption{Restricted $\color{our_red}\mathbb{H}_c$ in the exploration space of $\color{our_blue}\mathbb{R}$.}
    \label{fig:exploration}
    \end{subfigure}
    \begin{subfigure}{.74\textwidth}
    \centering
    \includegraphics[width=\linewidth]{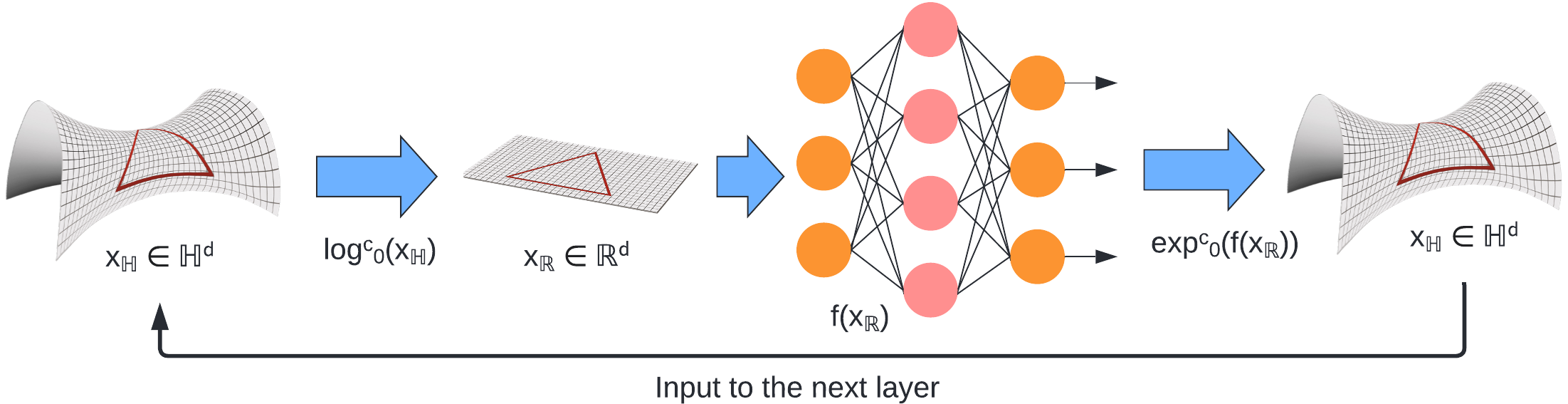}
    \vspace{-1em}
    \caption{Information loss due to frequent $\log_0^c$ and $\exp_0^c$ mapping to operate in the Euclidean space and represent in the hyperbolic space, respectively.}
    \label{fig:information_loss}
    \end{subfigure}
    \vspace{-.4em}
    \caption{Challenges in HNN learning. (a) shows the restricted domain of the hyperbolic space $\mathbb{H}_c$ with curvature $c$, in contrast to the actual exploration range of real numbers $\mathbb{R}$. HNNs clamp the gradients to avoid the out-of-domain errors which will lead to vanishing gradients in deep networks. (b) shows the information loss due to frequent mapping between the hyperbolic and Euclidean spaces. The loss is more pertinent in deeper networks due to the increased mapping frequency. }
    \label{fig:challenges}
    \vspace{-1em}
\end{figure}

Recent works on hyperbolic neural networks (HNNs) \cite{ganea2018hyperbolic} have shown that a number of problems in fields such as biology, network science, e-commerce, and computer vision possess a non-Euclidean anatomy \cite{bronstein2017geometric}. Consequently, this has led to the development of novel HNNs in the fields of graph analysis \cite{chami2019hyperbolic}, e-commerce \cite{choudhary2022anthem}, computer vision \cite{khrulkov2020hyperbolic}, and reasoning \cite{choudhary2021self}. These models, while superior to their Euclidean counterparts in performance, are only able to handle relatively smaller size datasets. The primary reason for this is their non-scalability on accelerated deep learning hardware such as GPUs and TPUs. The basic hyperbolic replacements of even simple operations (such as additions and multiplications) rely on computationally expensive and non-scalable hyperbolic functions (such as $\tanh, \cosh, \tanh^{-1}$, and $\cosh^{-1}$). Additionally, the gyrovector operations \cite{ganea2018hyperbolic} used in these models limit the hyperbolic vectors to the range of $\tanh^{-1}(x) \in (-1,1)$ (shown in Figure \ref{fig:exploration}). Note that here $c^{\frac{-1}{2}}$ is multiplied on both sides, where $c$ is the hyperbolic curvature. Due to this, the training of large hyperbolic models run into out-of-domain errors which are generally handled using domain clamping \cite{chami2019hyperbolic}. This leads to the issue of vanishing gradients in large HNNs. To tackle this problem, certain approaches \cite{guo2021free,nickel2017poincare} have proposed the idea of feature clipping to limit the radius of hyperbolic manifolds. 
Furthermore, as shown in Figure \ref{fig:information_loss}, a general HNN frequently uses the exponential and logarithmic maps to learn representations in the hyperbolic space and operate in the Euclidean space, respectively. While this approach is beneficial in adopting new Euclidean architectures to the hyperbolic space, it also leads to a significant loss of information. To avoid this issue, certain methods \cite{shimizu2021hyperbolic,bachmann2020constant} propose to formulate all HNN operations in the hyperbolic space. But, this increases the memory and time complexity of existing formulation, subsequently, reducing its scalability for minor improvements in performance. This approach also limits the adoption of research on Euclidean components such as optimization methods and activation functions in HNNs. 

To tackle these issues, we investigate the gyrovector formulation of HNNs in the Poincaré ball \cite{ganea2018hyperbolic} to determine their key benefits and implementation challenges. Our insights into the problem show that the primary advantage of HNNs is their ability to utilize the hyperbolic trigonometric functions and exponentially expand the representation space with increasing radius of the Poincaré ball, which also corroborates their impressive performance gains on hierarchical datasets. However, the hyperbolic trigonometry also leads to computational inefficiency due to GPU non-scalability \footnote{The underlying $\artanh$ and $\arcosh$ operations are not scalable on GPUs.}. Hence, \textit{we aim to retain the benefits of the hyperbolic representational space while mathematically reformulating them for effective computation}. To this end, we propose a mathematical approximation of the hyperbolic trigonometry functions using polynomial Taylor series expansions (PTSE) to define a computationally efficient set of HNN components. PTSE provides a pseudo-hyperbolic polynomial equivariant of the hyperbolic operations that is compatible with the current GPU hardware\footnote{polynomial operations only contain tensor additions and multiplications which are natively supported by GPUs.}, thus, enabling model parallelization and computing acceleration. Furthermore, given its Euclidean nature, the PTSE formulation can also utilize the advancements of Euclidean network components such as the ReLU activation \cite{nair2010rectified} to tackle the issue of vanishing gradients, and also does not require the frequent mapping between tangent and hyperbolic spaces. It should be noted that using PTSE instead of gyrovector formulation will lead to an approximation loss. However, this loss can be managed by tuning the precision of PTSE according to the dataset and hardware. In our empirical evaluations, we observe that PTSE formulations have 85\%  lower latency\footnote{latency $\propto \frac{1}{T}$, where $T$ is the total time taken to train the model.} than their gyrovector formulations. We also notice that the PTSE formulations are more scalable through empirical evidence of performance on multiple GPUs. Also, to evaluate the performance of PTSE formulation, we compare it against its hyperbolic and Euclidean counterparts on standard problems in the areas of graph analysis, natural language processing, and computer vision~\footnote{Note that we experiment with problems in three areas in order to avoid any biases to a particular dataset/domain.} The main contributions of this paper are as follows:
\begin{itemize}[noitemsep,leftmargin=*]
    \item We reformulate the hyperbolic network components by utilizing the Taylor series expansions of hyperbolic trigonometric functions for achieving scalability and compatibility with accelerated deep learning hardware. We handle the problem of approximation loss by tuning the PTSE precision. To the best of our knowledge, no other work has explored such an alternative formulation.
    \item We show that PTSE formulations, while maintaing comparable performance, are 85\% faster than their hyperbolic counterparts through empirical evaluation on standard problems in graph analysis and computer vision.
    \item We demonstrate the effect of precision and regularization factor in training the PTSE models through an ablation study. Also, we further show the inherent hyperbolic nature of PTSE formulation through experiments with both hyperbolic and Euclidean optimizers.
\end{itemize}

The rest of the paper is organized as follows; Section \ref{sec:related} describes the related research, Section \ref{sec:model} presents the hyperbolic operations with the proposed PTSE reformulation. Section \ref{sec:experiments} details our evaluation, Section \ref{sec:conclusion} concludes the paper and Section \ref{sec:broader_impact} discusses the broader impact of our work.

%% file: related.tex
\section{Related Work}
\label{sec:related}

\textbf{Hyperbolic Neural Networks.} HNNs have recently gained prominence in several research problems involving hierarchical datasets. For the popular Poincaré ball model with a curvature $c$, \citet{ganea2018hyperbolic} introduced the essential gyrovector operations for training neural networks such as Möbius addition ($\oplus_c$), Möbius scalar multiplication ($\odot_c$), parallel transport ($P_x^c$), exponential map ($\exp_x^c$), and logarithmic map ($\log_x^c$). Building on this, HGCN \cite{chami2019hyperbolic} and HypE \cite{choudhary2021self} models were developed for hyperbolic deep learning in graphs and knowledge graphs, respectively. In HAN \cite{gulcehre2018hyperbolic}, authors developed the Lorentz \cite{ungar2005analytic} gyrovector operations to model the attention mechanism for improvement on machine translation and visual question answering. HHAN \cite{zhang2021hype} and HGAN \cite{zhang2021hyperbolic} applied the framework for hierarchical aggregation and graph problems, respectively. In the e-commerce domain, ANTHEM \cite{choudhary2022anthem} showed an improved performance over the baselines for product search. In computer vision, Hyperbolic Protonet \cite{khrulkov2020hyperbolic} showed better performance over its Euclidean counterpart for image embeddings. While these methods show the impressive performance of HNNs, they are not scalable to practical problems with large datasets due to the underlying complexity of traditional HNN formulation.

\textbf{Hyperbolic Formulations.} In addition to the initial gyrovector operations \cite{ganea2018hyperbolic}, several new formulations have been developed to solve specific problems. Fully hyperbolic networks \cite{chen2021fully} and Hyperbolic networks++ \cite{shimizu2021hyperbolic} have formulated the gryovector operations completely in the hyperbolic space to avoid the frequent mapping from tangent space. These works also cover their respective formulation's extension to linear, convolution, and attention networks. \citet{mathieu2019continuous} constructs Poincaré variational autoencoders through Gaussian generalizations on the hyperbolic space. Another line of research \cite{guo2021free} provides a solution to vanishing gradients by limiting the radii of Poincaré ball model. While these methods are effective for their specific problems, they do not provide a complete solution to various limitations of a generalized hyperbolic learning model.

%% file: model.tex
\section{Taylor Series Approximation of Hyperbolic Operators}
\label{sec:model}
Our primary aim is to reformulate the hyperbolic gyrovector operations to (i) improve scalability, (ii) limit the problem of vanishing gradients and (iii) avoid the frequent mapping from the tangent space. We reformulate the essential gyrovector operations using Taylor series expansions with tunable precision and extend them to more complex neural network components such as layers, activation functions, and distance metrics.
\subsection{Taylor Series Reformulation of Gyrovector Operations}
For simplicity and ease of comparison across existing techniques, we show the reformulation of the popular Poincaré ball model of the hyperbolic space \cite{ganea2018hyperbolic}. However, the technique can be trivially extended to other hyperbolic space models such as the Beltrami-Klein \cite{sun2021hyperbolic} and Lorentz \cite{zhang2021lorentzian} model. The Poincaré ball formulation primarily relies on the hyperbolic trigonometric functions of $\tanh, \tanh^{-1}$, and $\cosh^{-1}$. However, these functions are not scalable on GPUs\footnote{GPUs are generally faster for deep learning due to their ability to parallelize basic operations such as tensor addition and multiplication. But, hyperbolic functions are complex and their GPU parallelization is non-trivial.} and have limited range which further leads to information loss. Hence, we approximate the behavior of these functions using their polynomial Taylor series expansion (PTSE).
\begin{definition}
\label{def:PTSE}
Hyperbolic trigonometric functions can be approximated with their PTSE of $n$ terms. The parameter $n$ is tunable to balance precision and computational costs, where higher value of $n$ implies lower approximation error but higher computational cost. The PTSE approximation $PTSE(f(x))$ and the relative approximation error $\eta$ are defined as follows.
\begin{flalign}
    \tanh(x) &\approx PTSE(\tanh(x)) = \sum_{i=1}^{n} \frac{(-1)^{i-1}2^{2i}(2^{2i}-1)B_nx^{2i-1}}{(2i)!}\\
    \artanh(x) &\approx PTSE(\artanh(x)) = \sum_{i=1}^{n} \frac{x^{2i-1}}{2i-1}\\
    \arcosh(x) &\approx PTSE(\arcosh(x)) = \sum_{i=1}^{n}\left(\frac{(-1)^{i+1}(2x-1)^i}{i}-\left(\frac{\prod_{j=1}^i\frac{2j-1}{2j}}{(2i)x^{2i}}\right)\right) \forall |x|\ge 1 \\
    \eta(f(x)) &= \left\|\frac{f(x)-PTSE(f(x))}{f(x)+\epsilon}\right\|;\quad \epsilon \ll f(x) 
\end{flalign}
Note that $B_n$ is the $n^{th}$ Bernoulli number and $\lim_{n \to \infty}\eta(f(x))=0$~\footnote{ Additional details on the approximation errors of the PTSE reformulation are provided in Appendix \ref{app:approximation}.}.
\end{definition}
Using the above definition, we reformulate the gyrovector space of the Poincaré ball of curvature $c$ by substituting the non-linear hyperbolic trigonometric functions with their PTSE equivariants. The original linear formulation of Möbius addition ($\oplus$) and  reformulated PTSE operations of exponential map ($\exp^\diamond_x$), logarithmic map ($\log^\diamond_x$), Möbius scalar multiplication ($\odot^\diamond$), Möbius matrix-vector product ($\otimes^\diamond$), and parallel transport ($P^\diamond_{0\rightarrow x}$) are as follows;
\begin{align}
    \label{eq:add} x\oplus y &= \frac{(1+2c\langle x,y\rangle+c\|y\|^2)x+(1-c\|x\|^2)y}{1+2c\langle x,y\rangle+c^2\|x\|^2\|y\|^2} \equiv K_xx+K_yy\\
    exp^\diamond_x(v) &= x \oplus \sum_{i=1}^{n} \frac{(-1)^{i-1}2^{2i}(2^{2i}-1)B_n\lambda_x^{2i-1}(\sqrt{c}\|v\|)^{2i-2}}{(2i)!}v \equiv K_xx+K_{\exp}v\\
    log^\diamond_x(y) &= \sum_{i=1}^{n} \frac{\lambda_x^{2i-1}(\sqrt{c}\|-x\oplus y\|)^{2i-2}}{2i-1} (-x\oplus y) \equiv K_xx+K_yy\\
    r \odot^\diamond x &= \sum_{i=1}^{n} \frac{(-1)^{i-1}2^{2i}(2^{2i}-1)B_n\left(\sqrt{c}\|\sum_{j=1}^{n} \frac{(\sqrt{c}\|x\|)^{2j-2}}{2j-1}rx\|\right)^{2i-2}}{(2i)!}. \frac{(\sqrt{c}\|x\|)^{2i-2}}{2i-1}rx\\
    &\equiv K_{rx}rx\\
    \label{eq:prod} M \otimes^\diamond v &= \sum_{i=1}^{n} \frac{(-1)^{i-1}2^{2i}(2^{2i}-1)B_n\left(\sum_{j=1}^{n} \frac{(\sqrt{c}\|v\|)^{2j-2}}{2j-1}\right)^{2i-1}}{(2i)!}Mv \equiv K_{Mv}Mv\\
    P^\diamond_{0\rightarrow x}(v) &= \sum_{i=1}^{n} \frac{\|m\|^{2i-1}}{2i-1}m \equiv K_xx+K_vv; ~~m = x \oplus \sum_{i=1}^{n} \frac{(-1)^{i-1}2^{2i}(2^{2i}-1)B_n(\sqrt{c}\|v\|)^{2i-2}}{(2i)!}v
\end{align}
In the above PTSE formulation, we note that the hyperbolic vector operations are reduced to their equivalent Euclidean operation scaled by polynomial functions\footnote{Note that polynomial functions are scalable on GPUs and are computationally faster to compute \cite{abdi2019gpu} than hyperbolic trigonometric functions.}, e.g., in $M \otimes^\diamond v = K_{Mv}Mv$, where $K_{Mv}$ is a polynomial function, $x \oplus y = K_xx+K_yy$, where $K_x$ and $K_y$ are polynomial functions. \textit{Thus, we are able to approximate and replace hyperbolic trigonometric functions, which are non-scalable on GPUs/TPUs, with a scalable equivalent that uses the scalable vector addition and multiplication operation scaled by polynomial functions for hyperbolic gyrovector operations.}

\subsection{Extension to Components of Hyperbolic Neural Networks}
In this section, we extend our PTSE gyrovector formulation to different HNN components including activation functions, layers, and optimizers. Note that, in the PTSE gyrovector formulation, we do not bound the operations in the hyperbolic space and hence the parameters of PTSE neural networks remain in the Euclidean vector space $\mathbb{R}^p$.

\textbf{Hyperbolic Activation.} We convert a Euclidean activation function $\varphi:\mathbb{R}^p \rightarrow \mathbb{R}^p$ to its PTSE hyperbolic equivalent $\varphi^\diamond:\mathbb{R}^p \rightarrow \mathbb{R}^p$ as follows;
\begin{equation}
    \nonumber\varphi^\diamond(x) = \sum_{i=1}^{n} \frac{(-1)^{i-1}2^{2i}(2^{2i}-1)B_n(\sigma\left(\|\sum_{i=1}^n\frac{\|\sqrt{c}x\|^{2i-2}}{2i-1}x\|\right))^{2i-2}}{(2i)!} \varphi\left(\sum_{i=1}^n\frac{\|\sqrt{c}x\|^{2i-2}}{2i-1}x\right)
\end{equation}
Here, $\varphi$ can be any Euclidean activation function including sigmoid, $\tanh$, and ReLU among others. 

\textbf{Hyperbolic Distance.} We define two distance metrics $d1_\diamond(x,y)$ and $d2_\diamond(x,y)$ in the PTSE gyrovector space as equivariants to the L1-norm and L2-norm in the Euclidean space, respectively. $d1_\diamond$ is efficient as a loss function in hyperbolic networks due to its higher sparsity in prediction tasks, whereas, $d2_\diamond$ is applied when vector similarity is the primary goal, e.g., query-keyword matching to calculate the relevance weights in attention models. For vectors $x,y \in \mathbb{R}^p$, the scalar PTSE distance metrics are defined as follows:
\begin{align}
    \nonumber d1_\diamond(x,y) &= \frac{2}{\sqrt{c}}\sum_{i=1}^{n} \frac{(\sqrt{c}\|-x\oplus y\|)^{2i-1}}{2i-1};\\
    \nonumber d2_\diamond(x,y) &= PTSE\left(arcosh\left(1 + 2\frac{\|x-y\|^2}{(1-\|x\|^2)(1-\|y\|^2)}\right)\right); 
\end{align}

\textbf{PTSE Linear Layer.} For input features $x \in \mathbb{R}^p$, the PTSE formulation of the hyperbolic linear layer $f^\diamond_{l}:\mathbb{R}^p\rightarrow\mathbb{R}^d$ is defined as $f^\diamond_l(x) = w\otimes^\diamond x \oplus b \equiv K_{wx}wx+K_bb \equiv K_{(wx,b)}(wx+b)$, where $w \in \mathbb{R}^{d\times p}$ is a learnable weight matrix and $b \in \mathbb{R}^d$ is the bias vector. Note that our formulation is able to handle homogeneous system of solutions by removing the constraint $\|wx\| \neq 0$, which is necessary in previous formulations \cite{ganea2018hyperbolic,chami2019hyperbolic}.

\textbf{PTSE GRU Layer.} We can also extend the linear layer to formulate the gates, reset gate ($r_t$) and update gate ($z_t$), of the hyperbolic GRU layer as follows;
\begin{equation}
    \nonumber r_t = \sigma^\diamond\left(W^r \otimes^\diamond h_{t-1} \oplus U^r \otimes^\diamond x_{t} \oplus b^r\right);\quad z_t= \sigma^\diamond\left(W^z\otimes^\diamond h_{t-1} \oplus U^z \otimes^\diamond x_{t} \oplus b^z\right); 
\end{equation}
where $\sigma^\diamond$ is the hyperbolic sigmoid activation function, $h_{t-1} \in \mathbb{R}^d$ is the output of the previous timestep, and $x_t \in \mathbb{R}^p$ is the input at current timestep. $W^r, W^z \in \mathbb{R}^{d \times d}$ are the learnable weight parameters of the reset and update gates, respectively for the previous timestep and $U^r, U^z \in \mathbb{R}^{d \times p}$ are the same for the input at current timestep. The formulation can be further simplified using the PTSE linear layer for computational efficiency \footnote{Details of the simplification and values of the polynomial functions are provided in Appendix \ref{app:simplification}.};
\begin{equation}
    \nonumber r_t \equiv \sigma^\diamond\left(K_{r_t}(W^rh_{t-1} + U^rx_{t} + b^r)\right);\quad z_t \equiv \sigma^\diamond\left(K_{z_t}(W^zh_{t-1} + U^zx_{t} + b^z)\right); 
\end{equation}
Finally, the reset memory $\tilde{h_t}$ and final output $h_t$ is calculated as follows;
\begin{equation}
    \nonumber \tilde{h_t} = \varphi^\diamond\left((Wdiag(r_t))\otimes^\diamond h_{t-1}\oplus U \otimes^\diamond x_t \oplus b\right);\quad
    h_t = h_{t-1} \oplus diag(z_t) \otimes^\diamond (-h_{t-1} \oplus \tilde{h_t})
\end{equation}
 where $\varphi$ is the appropriate hyperbolic activation function.

\textbf{PTSE Convolution Layer.} The hyperbolic convolution layer, as defined by \citeauthor{chami2019hyperbolic} \cite{chami2019hyperbolic} in the context of graph neural networks, consists of two sequential operations: feature transformation and aggregation. We formulate the PTSE feature transformation at the $l^{th}$ layer, $h_a^{\diamond,l}$, for input $x_a \in \mathbb{R}^p$ as a linear layer, i.e., $h_a^{\diamond,l} =  W^l\otimes^\diamond x_a \oplus b^l \equiv K_{(W^l,b^l)}(W^lx_a+b^l)$. The formulation of feature aggregation depends on the problem. In this section, we present the feature aggregation of convolution layers specifically in the context of hyperbolic graph neural networks that uses attention-based aggregation over the nodes' neighborhood. However, note that the solution can be trivially extended to other problems as well, with a change in aggregation methodology, e.g., for computer vision, one could use max-pooling based spatial aggregation. Given a node $a$ in graph $\mathcal{G}$ with a set of node neighbors $b \in \mathcal{N}(a)$), the PTSE feature aggregation at the $l^{th}$ layer is formulated as follows:
\begin{align}
    h_a^{\diamond,l} &= W^l\otimes^\diamond x_a^{l-1} \oplus b^l;\quad\textcolor{blue}{\text{Feature Transformation}}\\
    \nonumber w^l_{ab} &= \text{SOFTMAX}_{b \in \mathcal{N}(a)}\left(MLP\left(\sum_{i=1}^{n} \frac{(\sqrt{c}\|h_a^{\diamond,l}\|)^{2i-2}}{2i-1}h_a^{\diamond,l} \Bigg\Vert \sum_{i=1}^{n} \frac{(\sqrt{c}\|h_b^{\diamond,l}\|)^{2i-2}}{2i-1}h_b^{\diamond,l}\right)\right)\\
    \nonumber y^l_a &= \sum_{i=1}^{n} \frac{(-1)^{i-1}2^{2i}(2^{2i}-1)B_n(\sqrt{c}\|m\|)^{2i-1}}{(2i)!}m;\quad m = \sum_{b\in \mathcal{N}(a)}w^l_{ab}\sum_{i=1}^{n} \frac{(\sqrt{c}\|h_b^{\diamond,l}\|)^{2i-2}}{2i-1}h_b^{\diamond,l}\\
    x_a^{l} &= \sigma^\diamond\left(y^l_a\right);\quad\textcolor{blue}{\text{Feature Aggregation and input to next layer}}
\end{align}
where $W^l \in \mathbb{R}^{d\times p}, b^l \in \mathbb{R}^{p}, h_a^l \in \mathbb{R}^{p}$ are the weights, biases, and output of the feature transformation operation, $x_i^{l-1} \in \mathbb{R}^{p}$ is the input from previous layer, $w^l_{ab} \in \mathbb{R}^{d \times d}$ are the attention weights, $y_a^l \in \mathbb{R}^{d}$ is the encoding after feature aggregation, and $x_a^l \in \mathbb{R}^{d}$ is the output of PTSE convolution layer.

\textbf{PTSE Attention Layer.} The hyperbolic attention layer, as defined by \citeauthor{gulcehre2018hyperbolic} \cite{gulcehre2018hyperbolic}, involves the steps of hyperbolic matching and aggregation. The concept of matching in attention layer is meant to capture the similarity between the query and key. To capture this, we utilize $d2_\diamond$ as the distance measure for matching. Given $q_i,k_j, v_{ij} \in \mathbb{R}^p$ are the input query, key and value vectors, the output of the attention layer is calculated as follows: 
\begin{align}
    \alpha_{ij} &= \sigma^\diamond(-\beta d2_\diamond(q_i,k_j)-c); \quad \textcolor{blue}{\text{Calculate attention weights through PTSE matching.}}\\
    x_{i} &= \sum_j \left[\frac{\alpha_{ij}\gamma(v_{ij})}{\sum_l\alpha_{il}\gamma(v_{il})}\right]v_{ij};~~\gamma(v_{ij}) = \frac{1}{\sqrt{1-\|v_{ij}\|^2}};\quad \textcolor{blue}{\text{Scaled output given in \cite{gulcehre2018hyperbolic}.}}
\end{align}
where $\beta$ and $c$ are the temperature and bias parameters that control the scale/variance and counteract the negative values of the distance metric, respectively. $\gamma(v_{il})$ is the Lorentz factor \cite{gulcehre2018hyperbolic} used to normalize the outputs of the attention layer.

\begin{figure}[htbp]
\vspace{-.8em}
    \centering
    \begin{subfigure}{.3\textwidth}
    \centering
    \includegraphics[width=\linewidth]{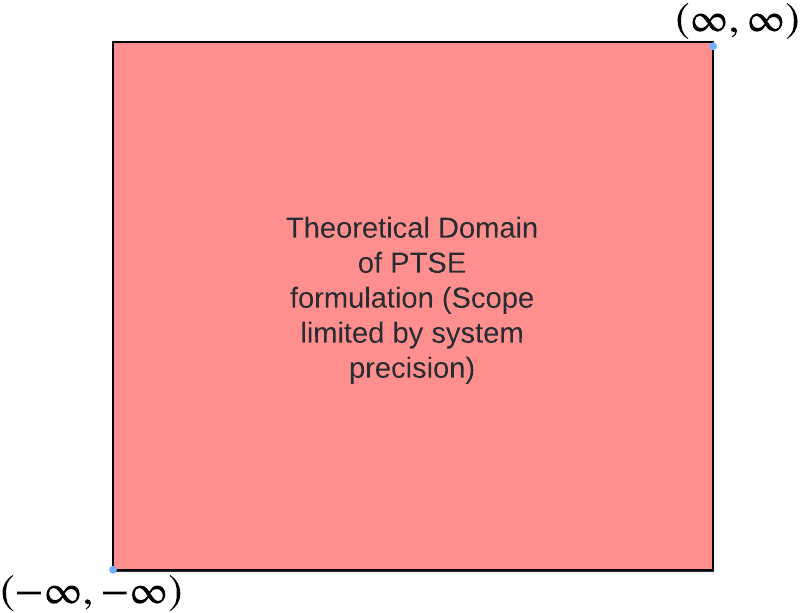}
    \vspace{-1em}
    \caption{PTSE formulation extends the exploration space to $\mathbb{R}$.}
    \label{fig:solve_exploration}
    \end{subfigure}
    \begin{subfigure}{.64\textwidth}
    \centering
    \includegraphics[width=\linewidth]{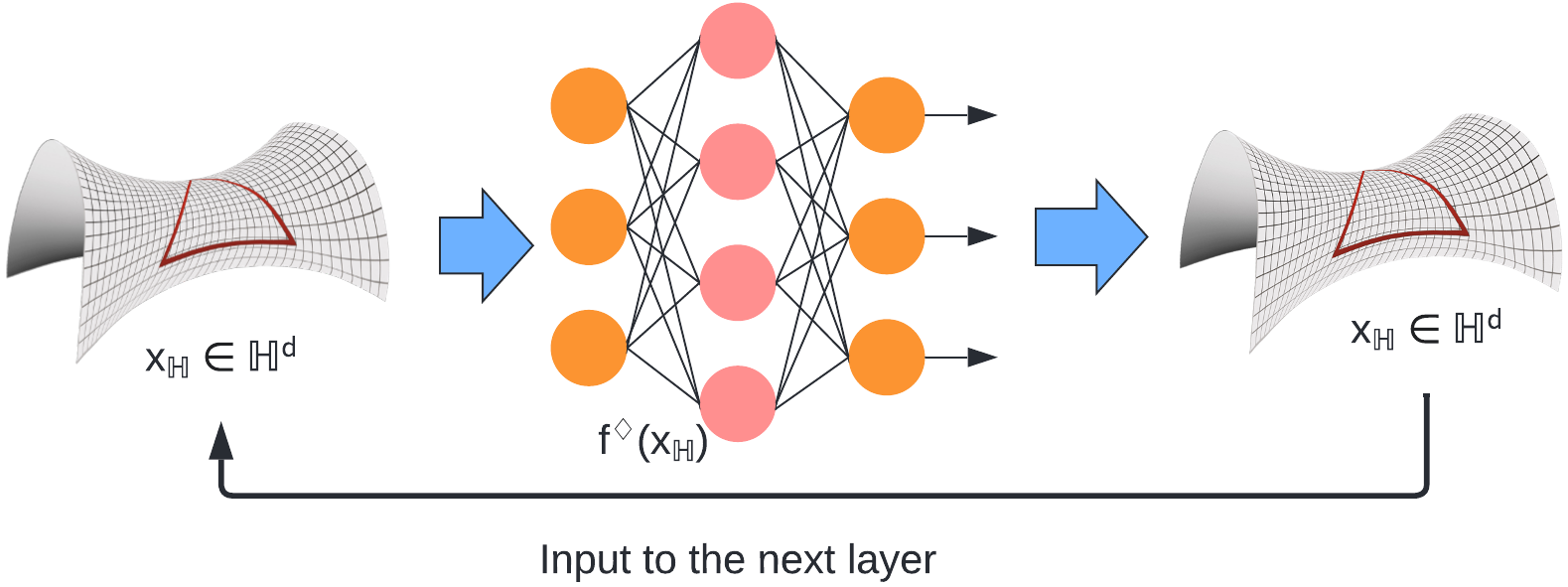}
    \vspace{-1em}
    \caption{PTSE formulations can directly be applied to embeddings in the hyperbolic space, thus avoiding a frequent back and forth mapping.}
    \label{fig:solve_information_loss}
    \end{subfigure}
    \vspace{-.4em}
    \caption{PTSE formulation for learning. (a) shows the extended domain of PTSE learning, i.e., the range of real numbers $\mathbb{R}$. PTSE does not need to clamp the gradients as there are no out-of-domain errors. (b) PTSE are able to operate on hyperbolic embeddings directly, thus, avoiding the mapping back and forth and limiting the information loss. }
    \label{fig:PTSE}
    \vspace{-1.2em}
\end{figure}
Thus, we have the PTSE formulations for the necessary layers, activation functions, and distance metrics which can be combined together in deep learning frameworks. Given their formulation, we see that the PTSE parameters are not restricted to $\mathbb{H}_c$ and can explore the entire Euclidean space $\mathbb{R}$ for solutions. Also, the PTSE layers do not need the frequent mapping between hyperbolic space and tangent space and can be computed in a single operation, i.e., $f^\diamond(x) = K_{f(x)}f(x)$. This tackles our major issues of closure (Figure \ref{fig:solve_exploration}) and loss of information (Figure \ref{fig:solve_information_loss}) in the hyperbolic space. In our experimental study, we use these formulations to construct certain hyperbolic models and compare them against previous baselines in their respective research domains.

%% file: experiments.tex
\section{Experimental Study and Evaluation}
\label{sec:experiments}
In this section, we evaluate our model by addressing the following research questions (RQs);
\begin{itemize}[noitemsep,leftmargin=*]
    \item \textbf{RQ1:} How do models based on PTSE formulation compare to the models utilizing the existing hyperbolic formulation \cite{ganea2018hyperbolic} and Euclidean formulation in terms of performance and computational complexity?
    \item \textbf{RQ2:} Can the PTSE formulation be extended to hyperbolic models in the domains of multi-relational reasoning, computer vision, and natural language processing?
    \item \textbf{RQ3:} How do the different components and design choice decisions affect PTSE model's performance?
\end{itemize}
\subsection{Problem Settings}
For the comparative analysis, we adopt the following problem settings for our experiments.

\textbf{Graph Prediction.} Given a graph $\mathcal{G} = (V \times E)$, with $v_i \in V$ are the set of vertices with labels $y_i$, and $e_{ij} \in E$ is the Boolean adjacency matrix, where $e_{ij} = 1$ if a link exists between $v_i$ and $v_j$, and $0$ otherwise. Based on this context, we formulate the problem of \textit{node classification}, where the aim is to estimate a predictor model $P_\theta$, parameterized by $\theta$, such that for an input node $v_i$, $P_\theta(v_i|E)=y_i$, and the problem of \textit{link prediction}, where for an input node-pair $v_i, v_j$ and incomplete adjacency matrix $E^{in}$, the objective to estimate a predictor model $P_\phi$, such that $P_\phi(v_i,v_j|E^{in})=e_{ij}$.

\textbf{Multirelational Reasoning.} Given a knowledge graph of triplets, ${h_i,r_{ij},t_j} \in \mathcal{KG}$, where $h_i,r_{ij},t_j \in \mathbb{R}^n$ are the head, relation and tail, respectively. In the task of \textit{reasoning}, the goal is to predict tails $t_j \in \mathcal{KG}$ such that they answer the query of head $h_i$ transformed by relation $r_{ij}$.

\textbf{Few-shot Image Classification.} Let us consider a dataset of images $x_i \in X$ with corresponding labels $y_i$. Additionally, we also have a backbone embedding model $E_\phi$ trained on a large corpus $C$, where $|X|\ll |C|$. The goal here is to estimate a predictor model $P_\theta$ in a limited number of training steps $n$ such that $P(x_i|X)=y_i$.

\textbf{Textual Entailment.} Given two sentences, premise $p_i$ and a hypothesis $h_i$, the goal here is to estimate a $P_\theta$ such that $P_\theta(p_i,h_i)=y_i$, where $y_i=1$ if $p_i$ entails $h_i$, and $y_i=0$, otherwise.

\textbf{Nosiy Prefix Detection.} Given two sentences, $s_i$ and $s_j$, the goal here is to estimate a $P_\theta$ such that $P_\theta(s_i,s_j)=y_{ij}$, where $y_{ij}=1$ if $s_i$ is a noisy prefix of $s_j$, or $y_{ij}=0$ if $s_i$ is just a random sentence.
\subsection{Datasets and Baselines}
The primary goal of our experimental study is to compare the underlying formulation criteria of different models. To this end, we select the standard datasets and benchmarks for hyperbolic networks in the areas of graph neural networks (GNN), multi-relational reasoning (MuR), computer vision (CV), and natural language processing (NLP). The datasets, baselines, and evaluation metrics used in our experiments are provided in Table \ref{tab:datasets}. Additional details are provided in Appendix \ref{app:datasets_baselines}.
\begin{table}[htbp!]
    \footnotesize
    \centering
    \vspace{-1.2em}
    \caption{Description of the datasets, baselines and evaluation metrics used in different experiments of our paper. To maintain fair comparison, the standard experimental setup is considered and the dataset splits, sizes, and evaluation metrics are the same ones that are given in the benchmark papers. The benchmarks for the domains of GNN, MuR, CV, and NLP are \cite{chami2019hyperbolic}, \cite{balazevic2019multi}, \cite{khrulkov2020hyperbolic}, and \cite{ganea2018hyperbolic}, respectively.}
    \resizebox{\textwidth}{!}{
    \begin{tabular}{l|l|l|l}
    \hline
    \textbf{Domain} & \textbf{Dataset} & \textbf{Our models and baselines} & \textbf{Evaluation Metrics}  \\\hline
    \textbf{GNN}  & Cora \cite{rossi2015the} & GCN \cite{kipf2017semi}, GAT \cite{velickovic2018graph} (Euc) & Accuracy (node classification)\\
           & Pubmed \cite{sen2008collective} & HGCN \cite{chami2019hyperbolic}, HGAT \cite{zhang2021hyperbolic} (Hyp) & Area under ROC (link prediction)\\
           & Citeseer \cite{sen2008collective} & TGCN, TGAT (PTSE, ours)\\\hline
    \textbf{MuR}    & FB15k \cite{bordes2013translating,toutanova2015representing} & MuRE \cite{balazevic2019multi} (Euc) & HITS@K \\
           & WN18RR \cite{dettmers2018convolutional}& MuRP \cite{balazevic2019multi} (Hyp)& (to compare precision in top-K results)\\
           &       & MuRT (PTSE, ours) & Mean Reciprocal Rank (to compare ranking)\\\hline
    \textbf{CV}     & MiniImageNet \cite{russakovsky2015imagenet} & ProtoNet \cite{snell2017prototypical} (Euc)\\
           & Caltech- & H-ProtoNet \cite{khrulkov2020hyperbolic} (Hyp)& Classification Accuracy\\
           & UCSD-Birds \cite{triantafillou2017few}       & T-ProtoNet (PTSE, ours)\\\hline
    \textbf{NLP}    & SNLI \cite{bowman2015large}& RNN \cite{mikolov2010recurrent}, GRU \cite{junyoung2014empirical}  (Euc)\\
           & PREFIX-K\% \cite{ganea2018hyperbolic}&  HRNN \cite{ganea2018hyperbolic}, HGRU \cite{ganea2018hyperbolic} (Hyp) & Prediction Accuracy\\
           &           &  TRNN, TGRU (PTSE, ours)\\
    \hline
    \end{tabular}}
    \label{tab:datasets}
    \vspace{-1em}
\end{table}

\textbf{Implementation Details.} We conducted our experimental study on a multi-GPU setup of Quadro RTX 8000 with 48 GB of VRAM each. While our systems did have 48 GB of RAM, a reproduction of our experiments only requires around 12 GB and the actual requirement is much lower ($\approx 8$ GB) for PTSE formulation. All our models and corresponding baselines are implemented in PyTorch \cite{paszke2019pytorch} to avoid cross-framework inaccuracies in comparison. The graph experiments also utilize the GraphZoo \cite{vyas2022graphzoo} framework. For the experiments involving computational complexity, the models have been run in isolation to remove any overhead due to memory management processes. Our models and corresponding baselines have all been tuned for best performance. An interesting point of note is on the optimizer choice between Riemannian Adam (RAdam) \cite{becigneul2018riemannian} and Euclidean Adam \cite{Kingma2015Adam}. As expected, RAdam generally works better for HNNs and Adam for Euclidean NNs, but it is interesting to see that RAdam also works better for PTSE formulation, although the formulation is in the Euclidean space. This provides certain evidence of the inherent hyperbolic nature of PTSE formulation, even though the formulation lies in the Euclidean space. Another important pratical detail is the use of weight regularization for PTSE layers. In the PTSE approximation errors, provided in Appendix \ref{app:approximation}, we observe that $\eta(f(x)) \propto \|f(x)\|$ and hence to maintain low values of $f(x)$, we add an L1-regularization factor to our layers $\lambda \|f(x)\|$, where $\lambda > 0$, i.e., final output of our layers $f^\diamond_{fin}(x) = f^\diamond(x) + \lambda \|f(x)\|$. In this case, $\lambda$ works as a penalizing control that depends on the nature of $f(x)$ and the corresponding precision $n$, e.g., functions like $arcosh(x)$ need lower $\lambda$, whereas, $artanh(x)$ need higher $\lambda$, and also, low values of $n$ need higher $\lambda$ and vice-versa \footnote{The code of our models and the baselines are provided at \url{https://github.com/Akirato/TSE-hyperbolic}}.

\subsection{RQ1: Comparative Analysis of the PTSE formulation}
To evaluate the performance of the PTSE formulation, we compare it against the Euclidean and hyperbolic formulation in the well-studied area of graph neural networks. We analyze the formulations on the problems of node classification and link prediction and utilize the standard benchmarks and evaluation criteria, as provided in Table \ref{tab:datasets}.
\begin{table}[htbp!]
    \centering
    \caption{Performance comparison results between PTSE (ours) and the baselines on the tasks of node classification and link prediction in the graph domain. The columns present the evaluation metrics of Accuracy and ROC for node classification and link prediction, respectively. The cells with the best performance are highlighted in bold and the second-best performance is underlined.}
    \begin{tabular}{l|ccc|ccc}
    \hline
         & \multicolumn{3}{c|}{\textbf{Node Classification Accuracy (\%)}}& \multicolumn{3}{c}{\textbf{Link Prediction ROC in (\%)}}\\
        \textbf{Models} & \textbf{Cora} & \textbf{Pubmed} & \textbf{Citeseer} & \textbf{Cora} & \textbf{Pubmed} & \textbf{Citeseer}\\\hline
        \textbf{GCN} & 80.1&78.5&\underline{71.4}&90.2&92.6&91.3\\
        \textbf{GAT} & 82.7&79.0&\textbf{71.6}&89.6&92.4&\textbf{93.6}\\\hline
        \textbf{HGCN} & 77.9&78.9&69.6&\underline{91.4}&\textbf{95.0}&\underline{92.8}\\
        \textbf{HGAT} & \underline{79.6}&79.2&68.1&90.8&93.9&92.2\\\hline
        \textbf{TGCN} & 77.6 & \underline{80.8}& 68.8 & \textbf{91.6} & 94.2 & 92.3\\
        \textbf{TGAT} & \textbf{81.7} & \textbf{81.3} & 69.9 & 91.2 & \underline{94.4}	& 92.7 \\\hline
    \end{tabular}
    \label{tab:gnn_domain}
\end{table}

\textbf{Performance Comparison.} The comparative results provided on the tasks of node classification and link prediction show that, despite the approximation errors, PTSE formulation is able to perform similar to the standard hyperbolic networks. In the case of node classification, we observe that TGAT is able to outperform the baselines, which is due to the extended exploration space of the solutions. In the case of link prediction, where message propagation is the primary focus, we observe that the performance of PTSE formulation and hyperbolic networks is similar, and hence, we conclude that approximation errors are not a limiting factor for performance.

\begin{figure}[htbp!]
    \centering
\begin{subfigure}{.24\textwidth}
    \centering
    \includegraphics[width=\textwidth]{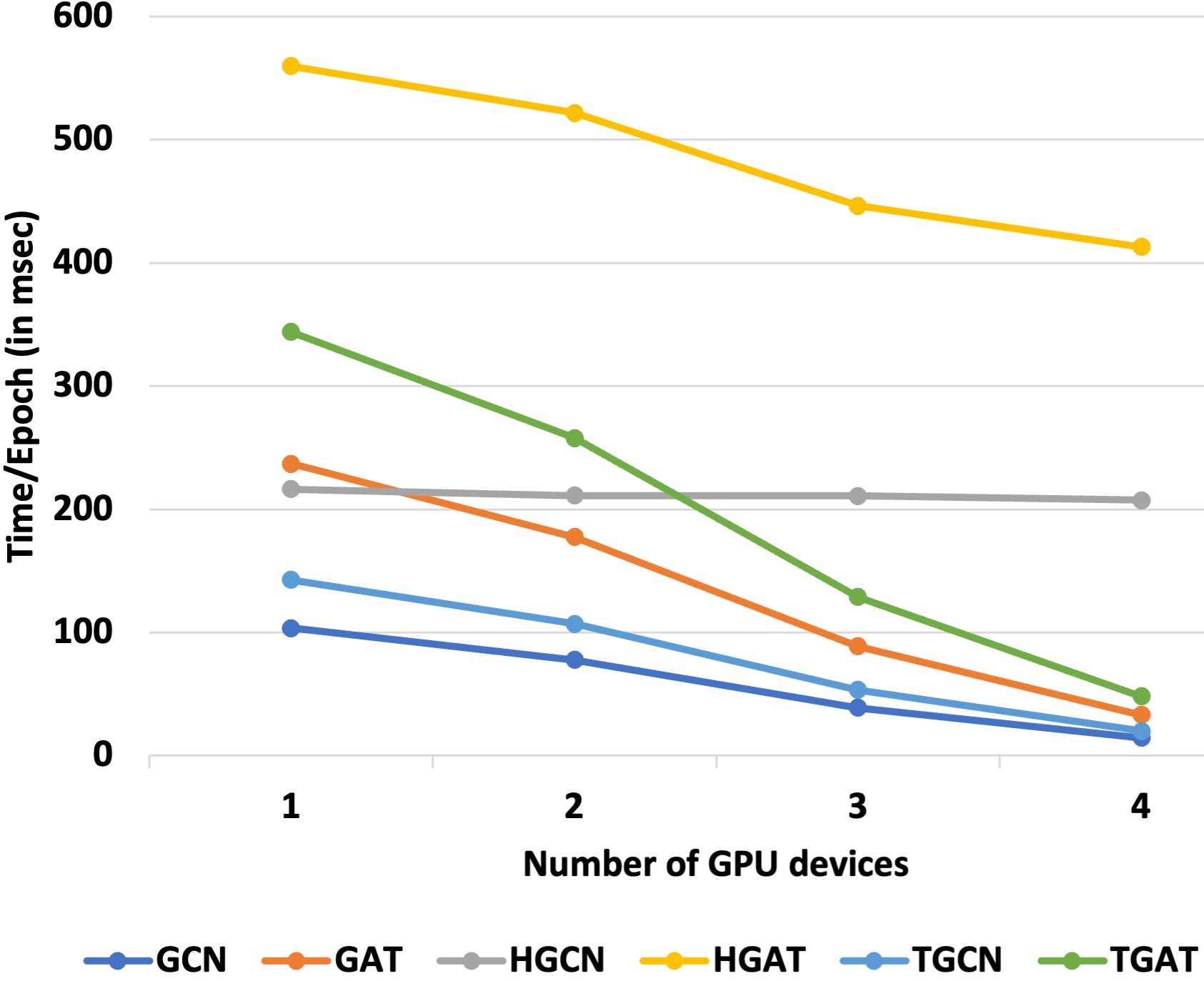}
    \caption{Time/Epoch (vs) \# GPU for Node classification.}
\end{subfigure}\hfill
\begin{subfigure}{.24\textwidth}
    \centering
    \includegraphics[width=\textwidth]{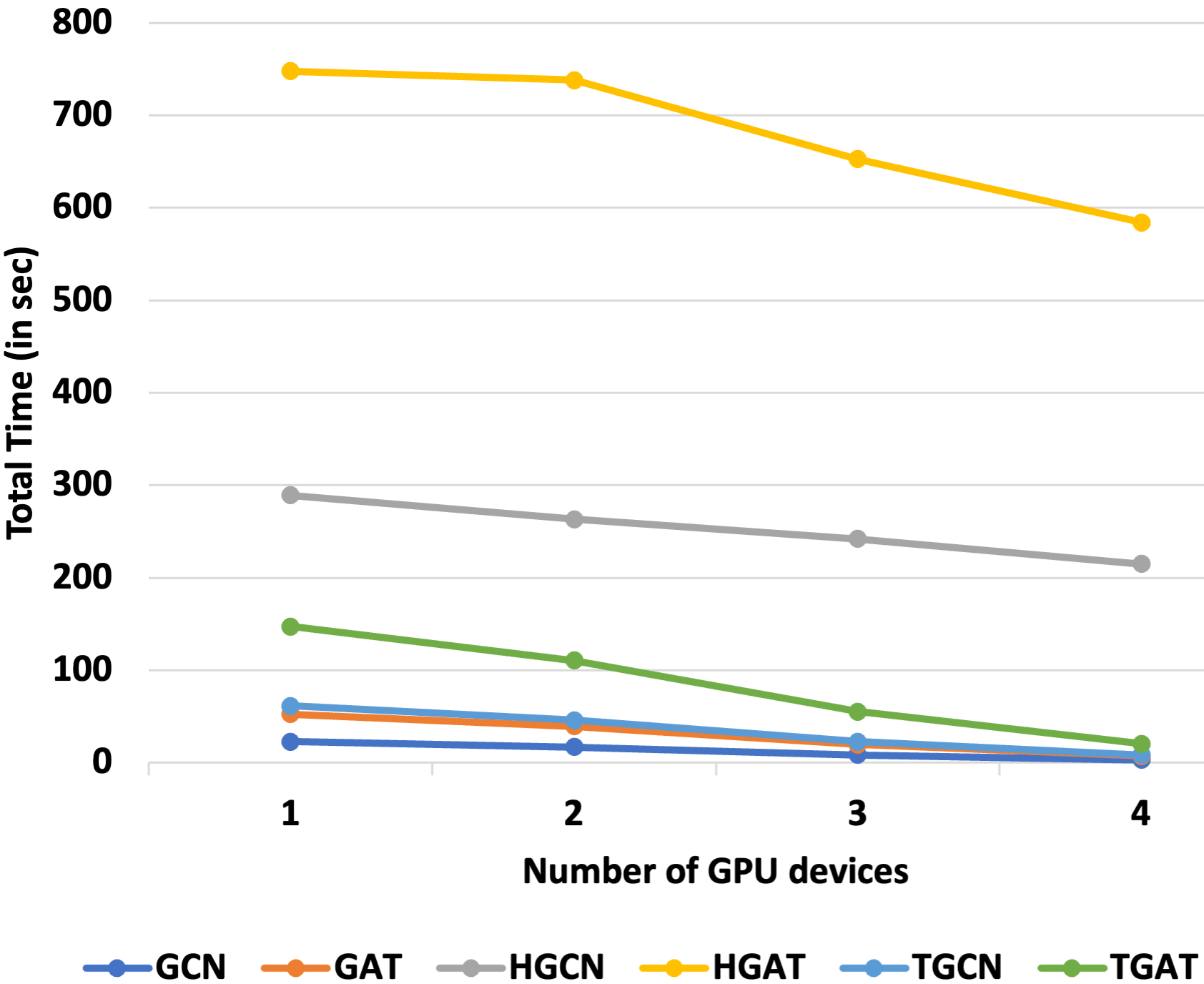}
        \caption{Total time (vs) \# GPU for Node classification.}
\end{subfigure}
\begin{subfigure}{.24\textwidth}
    \centering
    \includegraphics[width=\textwidth]{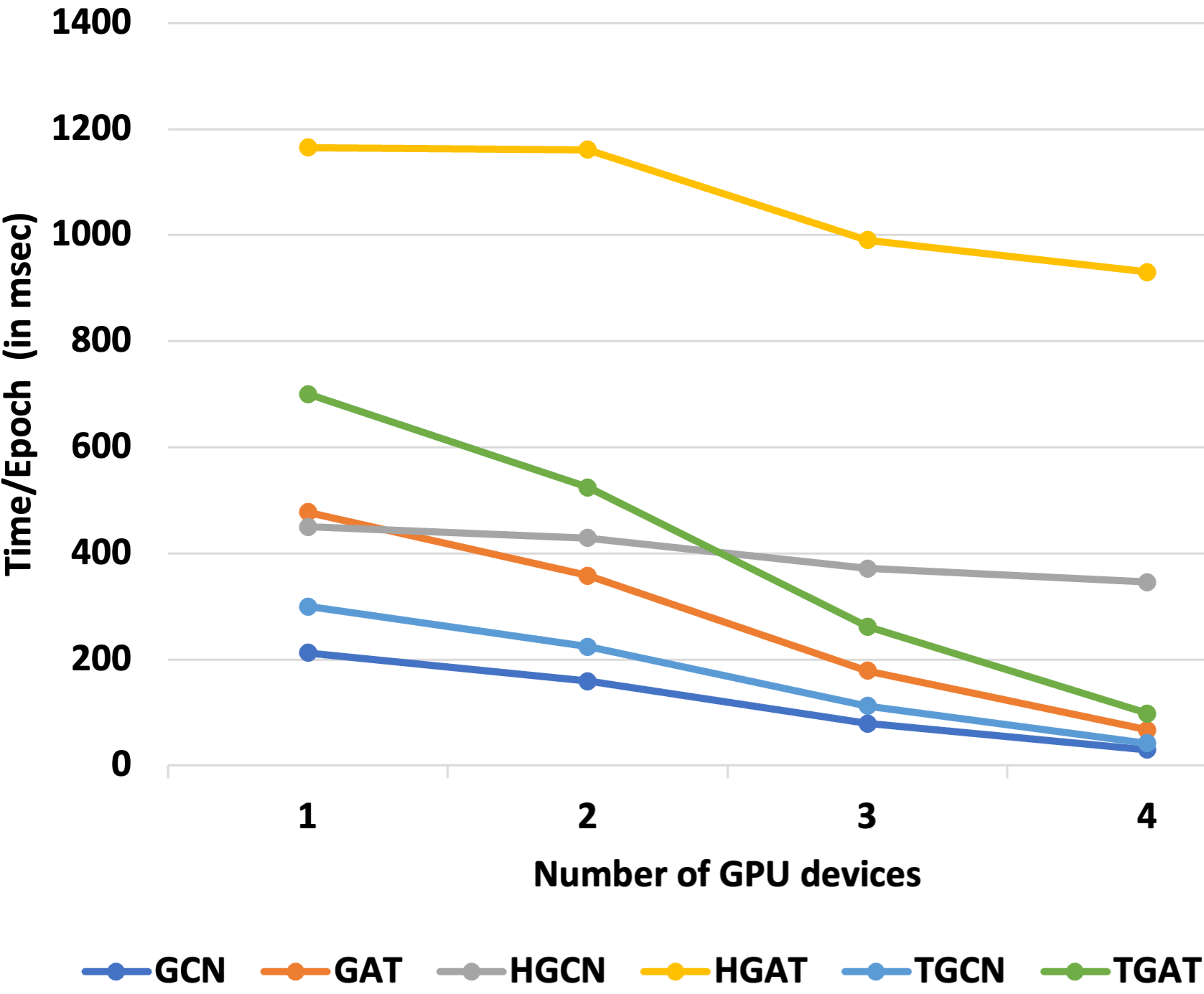}
        \caption{Time/Epoch (vs) \# GPU for Link prediction.}
\end{subfigure}\hfill
\begin{subfigure}{.24\textwidth}
    \centering
    \includegraphics[width=\textwidth]{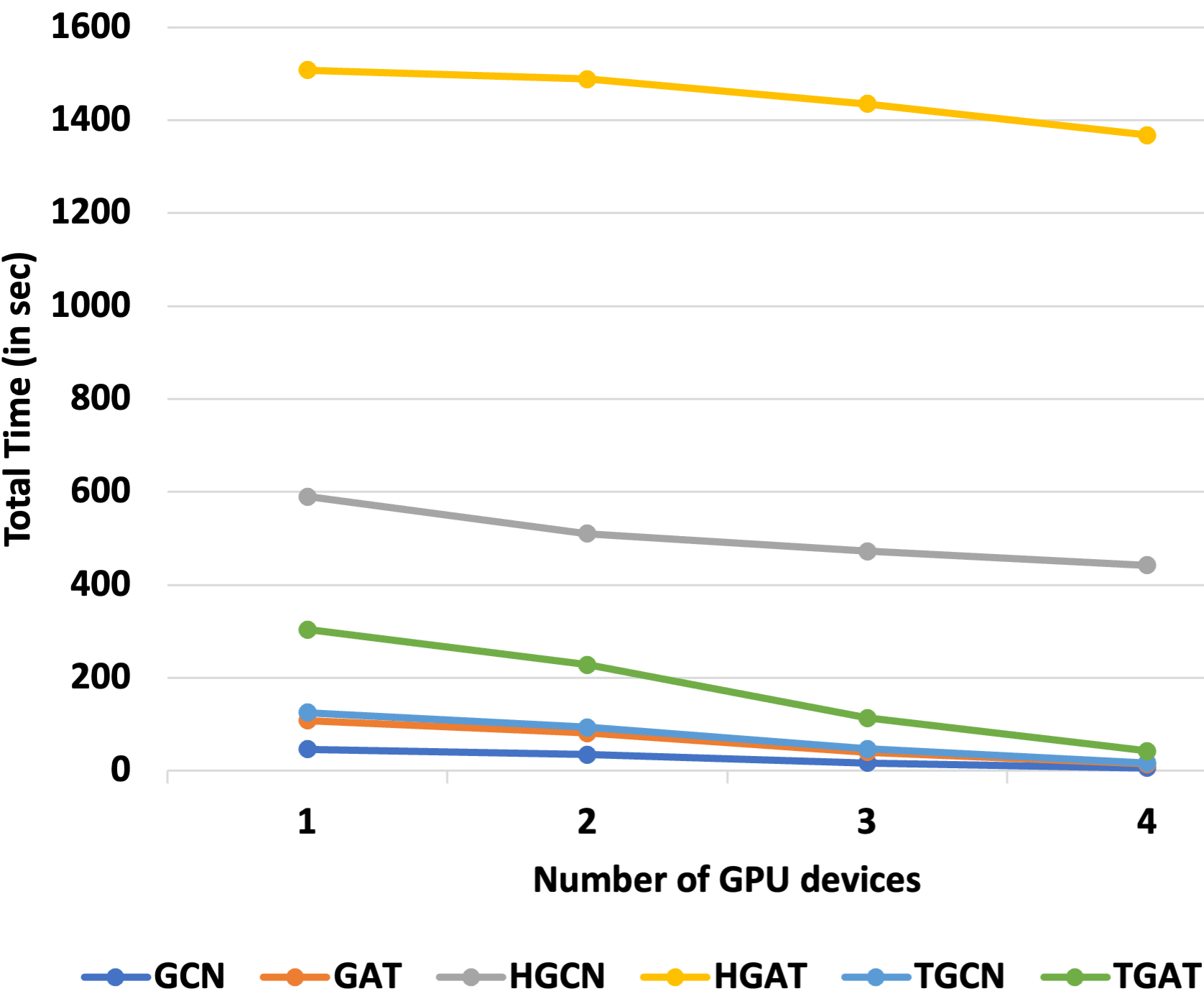}
        \caption{Total time (vs) \# GPU for Link prediction.}
\end{subfigure}%
\caption{Speed-up by parallelizing the process over multiple GPUs. Both Euclidean (GCN, GAT) and PTSE (TGCN, TGAT) formulations are able to leverage the multi-GPU setup and reduce the time per epoch and total time taken for convergence in both node classification and link prediction. We observe that, although both HNN and PTSE formulations are slower than Euclidean models on a single-GPU setup, PTSE formulations parallelize over multiple GPUs, to maintain their high performance while lowering their latency to similar times as Euclidean models.}
\label{fig:speed-up}

\end{figure}
\textbf{Computational Complexity.} From the results shown in Figure \ref{fig:speed-up}, we observe that both Euclidean and PTSE formulations are significantly faster and are able to speed-up when parallelized over multiple GPUs, in terms of both the time taken per epoch and the total time taken for model convergence. We also notice a slight dip in the time taken by hyperbolic methods, which is due to the parallelization of certain Euclidean operations in the HNNs. However, given the insignificant amount of dip for hyperbolic networks, we conclude that the PTSE formulation is significantly more parallelizable than its hyperbolic equivalent. From the comparative results on performance and computational complexity, we note that PTSE formulation, while similar in performance to hyperbolic networks, is much more scalable on deep learning hardware (GPUs in this case).

\subsection{RQ2: Case study of PTSE applications}
In this section, we study the implications of using the PTSE formulations in the domains of multi-relational reasoning, computer vision, and natural language processing. This study helps us understand the variety of applications that could leverage PTSE formulations.

\begin{table*}[htbp]
\centering
\caption{
Performance comparison results between MuRT (ours) and the baselines on the task of multi-relational graph reasoning in the graph domain. The columns present the evaluation metrics of Hits@K (H@K) (\%) and Mean Reciprocal Rank (MRR) (\%). Best performing values are highlighted in bold and the second best performing ones are underlined.}
    \begin{tabular}{ll|cccc|cccc}
    \hline
&\textbf{Datasets}&\multicolumn{4}{c|}{\textbf{WN18RR}}&\multicolumn{4}{c}{\textbf{FB15K-237}}\\\hline
\textbf{Dim}&\textbf{Models}&\textbf{MRR}&\textbf{H@10}&\textbf{H@3}&\textbf{H@1}&\textbf{MRR}&\textbf{H@10}&\textbf{H@3}&\textbf{H@1}\\\hline
\textbf{40}&\textbf{MuRE}&40.9&49.7&44.5&35.2&29.0&46.2&31.9&20.3\\
&\textbf{MuRP}&\textbf{42.5}&\underline{52.2}&\textbf{45.9}&\underline{36.3}&\textbf{29.8}&\textbf{47.4}&\textbf{32.8}&\underline{21.0}\\
&\textbf{MuRT}&\underline{42.2}&\textbf{52.6}&\underline{44.2}&\textbf{38.3}&\underline{29.6}&\underline{46.0}&\underline{32.6}&\textbf{22.2}\\\hline
\textbf{200}&\textbf{MuRE}&44.0&51.3&45.5&40.5&31.4&49.5&34.5&22.3
\\
&\textbf{MuRP}&\textbf{44.6}&\underline{52.4}&\textbf{46.2}&\underline{40.9}&\textbf{31.5}&\textbf{49.8}&\textbf{34.8}&\underline{22.5}\\
&\textbf{MuRT}&\underline{44.5}&\textbf{52.8}&\underline{46.0}&\textbf{41.0}&\underline{31.4}&\underline{49.6}&\underline{35.2}&\textbf{22.6}
\\\hline
    \end{tabular}
    \label{tab:multi-relation}
\end{table*}

\textbf{Multi-relational Reasoning.} In the case of multi-relational reasoning, the goal is to retrieve and rank a set of candidates, based on their relevance to a query head entity and the corresponding relation. We compare our methods using the standard metrics of HITS@K, which determines the precision of the models in their top-K results, and MRR, which quantifies the ranking ability of the models. In our empirical results (presented in Table \ref{tab:multi-relation}), we observe that our model MuRT is always able to provide results similar to its hyperbolic counterpart MuRP, but is significantly faster and more scalable than MuRP. Results in the areas of computer vision and natural language processing follow a similar trend and are presented in Appendix \ref{app:case-study}.

\begin{figure}[htbp!]
    \centering
\vspace{-1em}
\begin{subfigure}[b]{.3\textwidth}
    \centering
    \includegraphics[width=\textwidth]{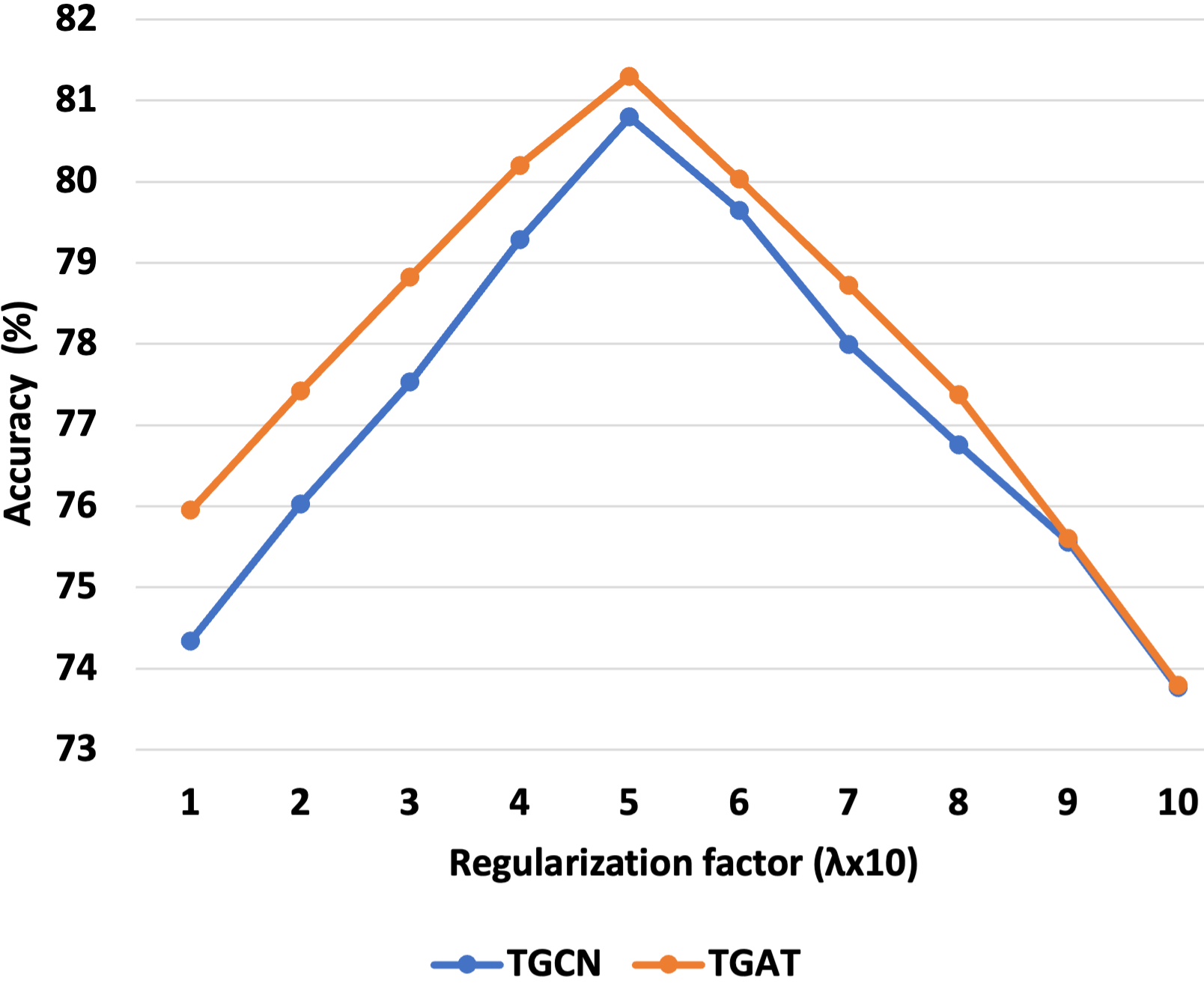}
    \caption{Regularization factor ($\lambda$) (vs) Accuracy.}
\end{subfigure}\hfill
\begin{subfigure}[b]{.3\textwidth}
    \centering
    \includegraphics[width=\textwidth]{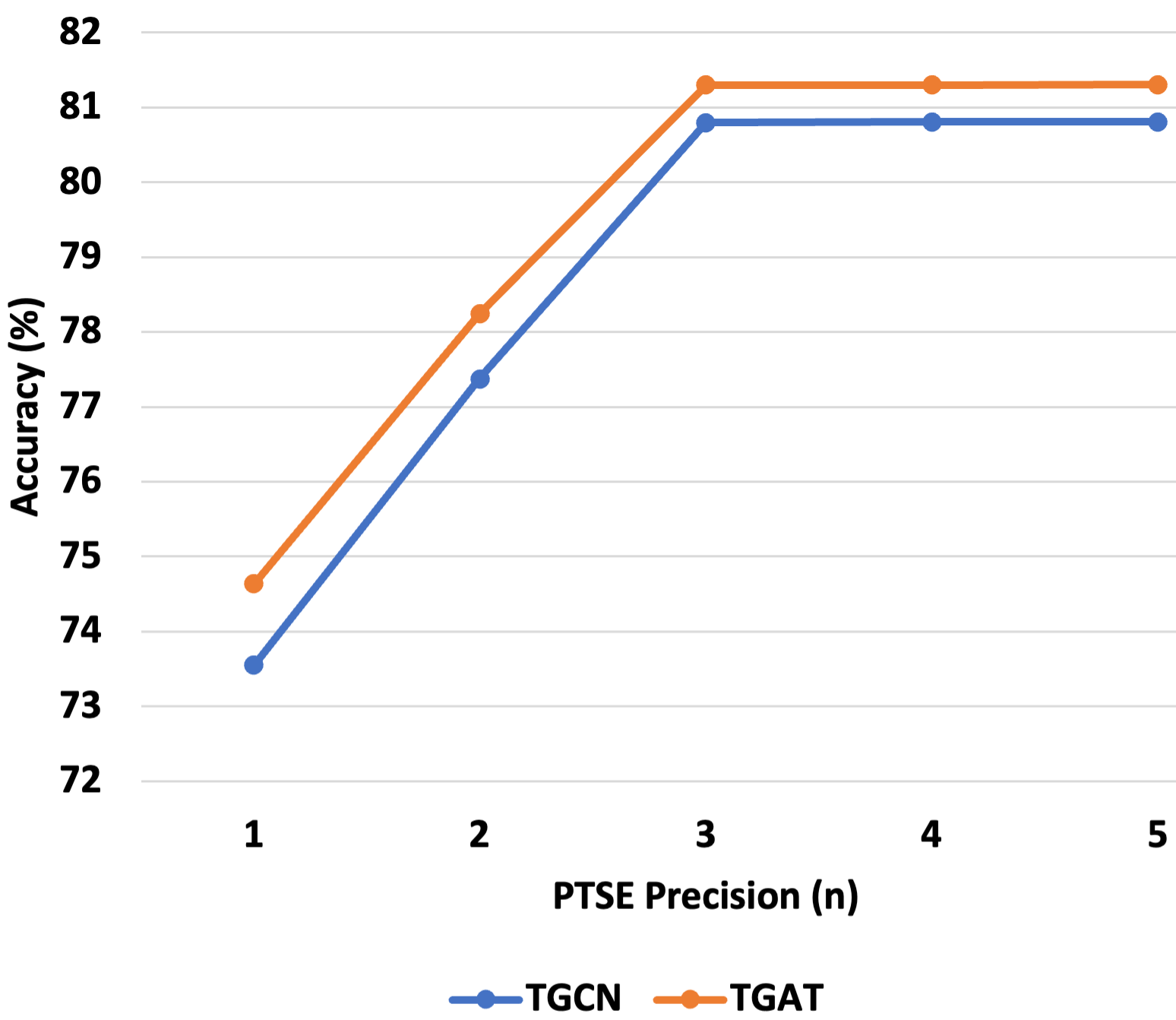}
        \caption{\footnotesize{PTSE Precision (n) (vs) Accuracy.}}
\end{subfigure}\hfill
\begin{subfigure}[b]{.35\textwidth}
    \centering
    \includegraphics[width=\textwidth]{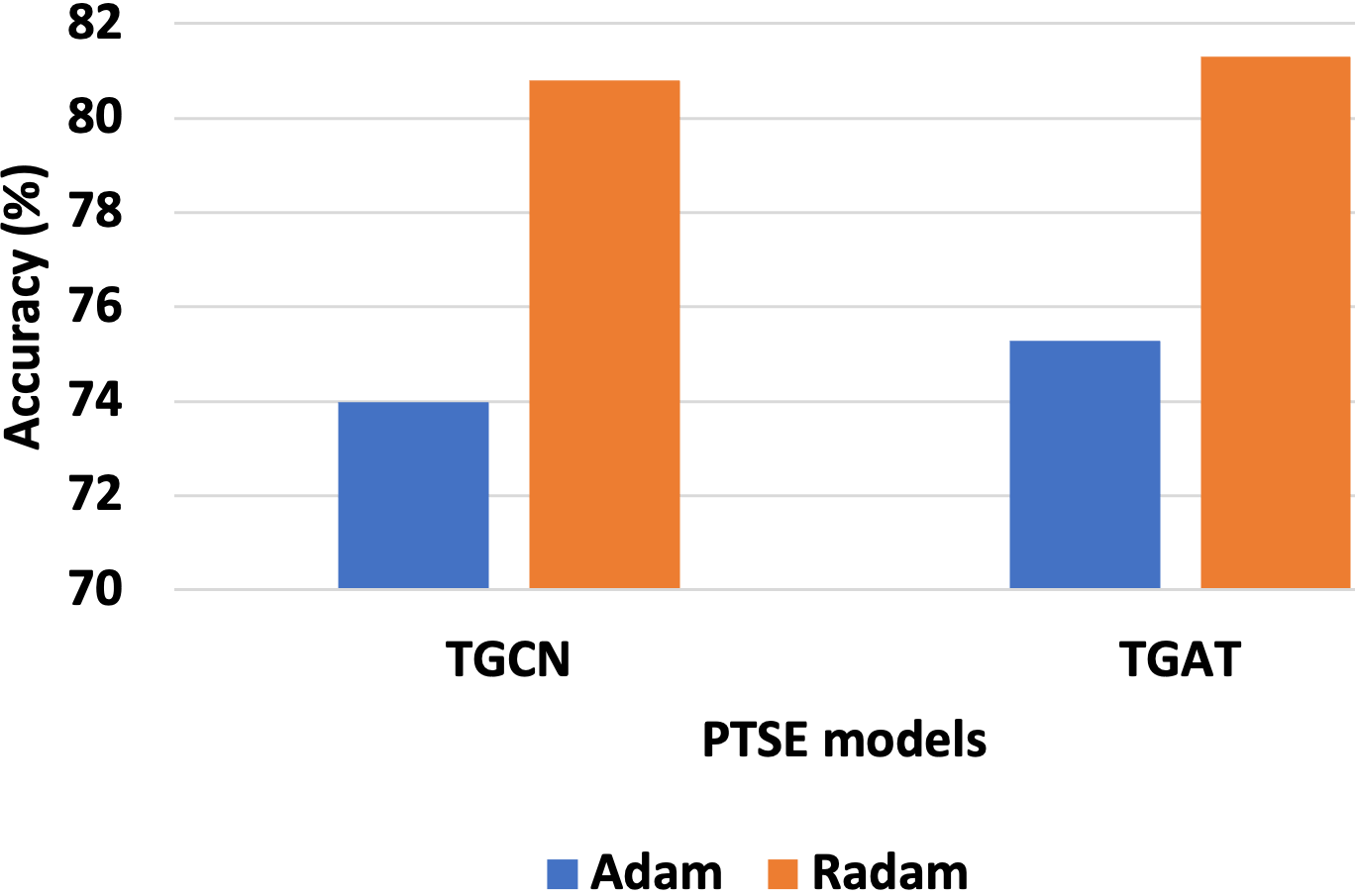}
        \caption{\footnotesize Choice of Optimizer.}
\end{subfigure}
\vspace{-0.6em}
\caption{Dependence of the performance of PTSE model (for node classification in the Pubmed dataset) on the following components: (a) regularization factor, (b) PTSE precision, and (c) Optimizer.}
\vspace{-1em}
\label{fig:components}
\end{figure}
\subsection{RQ3: Components of PTSE model}
This experiment aims to analyze the components of PTSE model and understand the effect of different design choices including various values of precision $n$, optimizers (Adam vs RAdam), and PTSE-regularization, to the overall performance of PTSE models.
As depicted in Figure \ref{fig:components}, we observe that Accuracy is dependent on several design choices. We note that the regularization factor is necessary to appropriately weigh the penalty of high values, as both a high and low value of $\lambda$ decreases the model performance. Additionally, we note that the hyper-parameter precision works as expected, where higher precision always returns a better accuracy due to low approximation error, however, the advantage is negligible after a certain value (n=3 in our case).  Furthermore, we also note that PTSE models perform better with RAdam optimizer, which provides further evidence of the inherent hyperbolic nature of the PTSE formulation.

%% file: conclusion.tex
\section{Conclusion}
\label{sec:conclusion}
In this paper, we present the PTSE formulation of the hyperbolic space and show that it is more scalable on GPUs/TPUs, compared to the original hyperbolic formulation. The primary reason for PTSE's effectiveness is its substitution of non-scalable tangent hyperbolic functions with more efficient Taylor series expansions. Additionally, we also note that PTSE formulation avoids the information loss due to frequent mapping back and forth between the tangent space and hyperbolic space, and also, extends the explorability of the solution space from the range of hyperbolic vectors $\mathbb{H}_c$ to the range of all real vectors $\mathbb{R}$. This allows us to apply the benefits of hyperbolic networks on a wider range of applications, especially problems that require the use of large-scale datasets.

%% file: broader_impact.tex
\section{Broader Impact}
\label{sec:broader_impact}
The problem of scalability has remained a primary challenge in the development of HNNs, particularly in their applicability to practical problems. Our PTSE formulation is a significant step forward in that direction, and provides evidence that approximate solutions can emulate the gains of hyperbolic networks in a scalable framework. However, certain understanding of the dataset is required for smooth application of the PTSE formulation. One needs to identify the hierarchical nature of the underlying dataset (hyperbolicity can be used as a proxy), as non-hierarchical datasets would benefit more from Euclidean architectures. Furthermore, our method adds the tunable hyper-parameters of precision and $\lambda$, that needs to be set appropriately, in order to achieve the best performance. However, certain preliminary experiments and an idea of the system precision could help in limiting the hyper-parameter space, thus reducing the effort to compute the precision and $\lambda$ parameter. Improving the scalability of hyperbolic neural networks could also improve the collection of data from graph datasets, thus, improving the efficiency of bad faith actors in collecting private information or attacking existing information storage systems.

%% file: appendix.tex
\appendix

\vspace{0.1in}
\section*{Appendix}

This document is an appendix to the paper titled ``Towards Practical Hyperbolic Neural Networks using Taylor Series Approximations''.
\section{Approximation Error of PTSE Expansions}
\label{app:approximation}
\begin{figure}[htbp!]
    \centering
\begin{subfigure}{.49\textwidth}
    \centering
    \includegraphics[width=\textwidth]{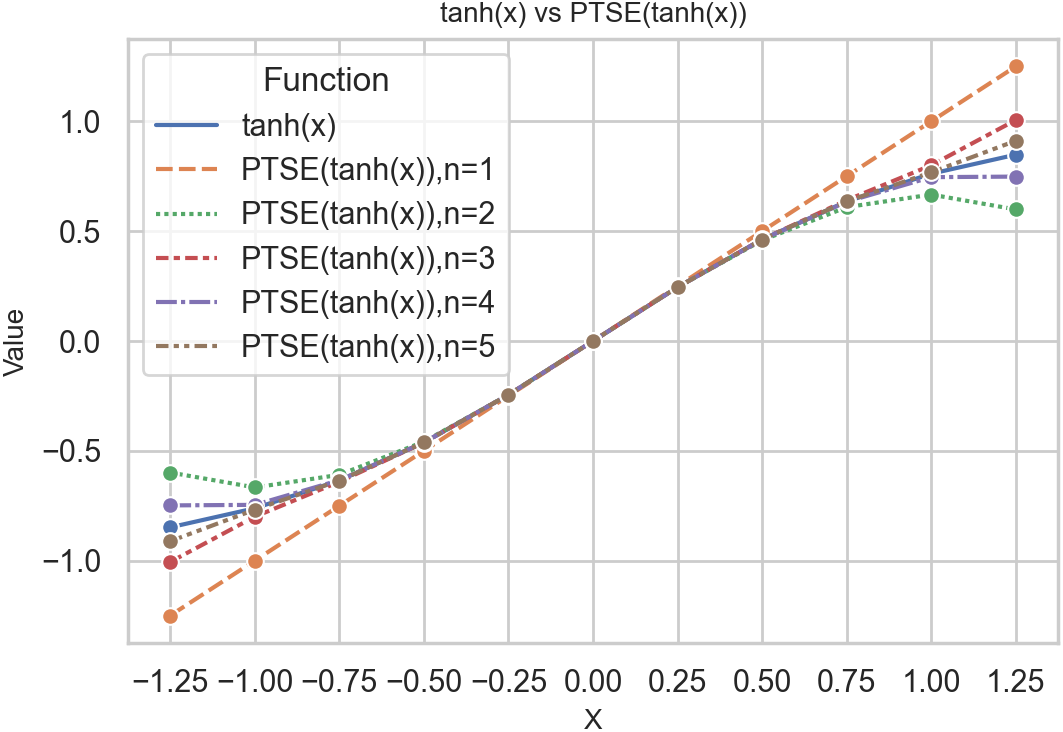}
    \caption{Plot of the $\tanh$ (solid line) function and its PTSE formulation at different values of precision.}
\end{subfigure}\hfill
\begin{subfigure}{.49\textwidth}
    \centering
    \includegraphics[width=\textwidth]{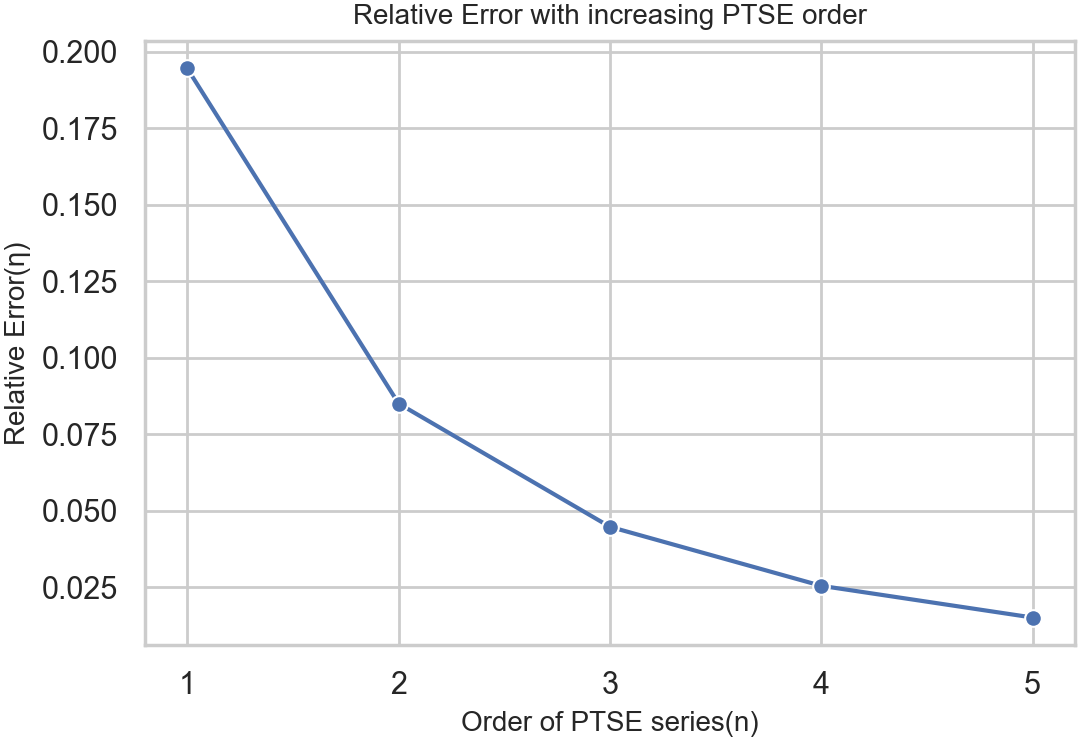}
        \caption{Relative Error between $\tanh$ and the PTSE formulation at different values of precision ($n$ vs $\eta$).}
\end{subfigure}
\begin{subfigure}{.49\textwidth}
    \centering
    \includegraphics[width=\textwidth]{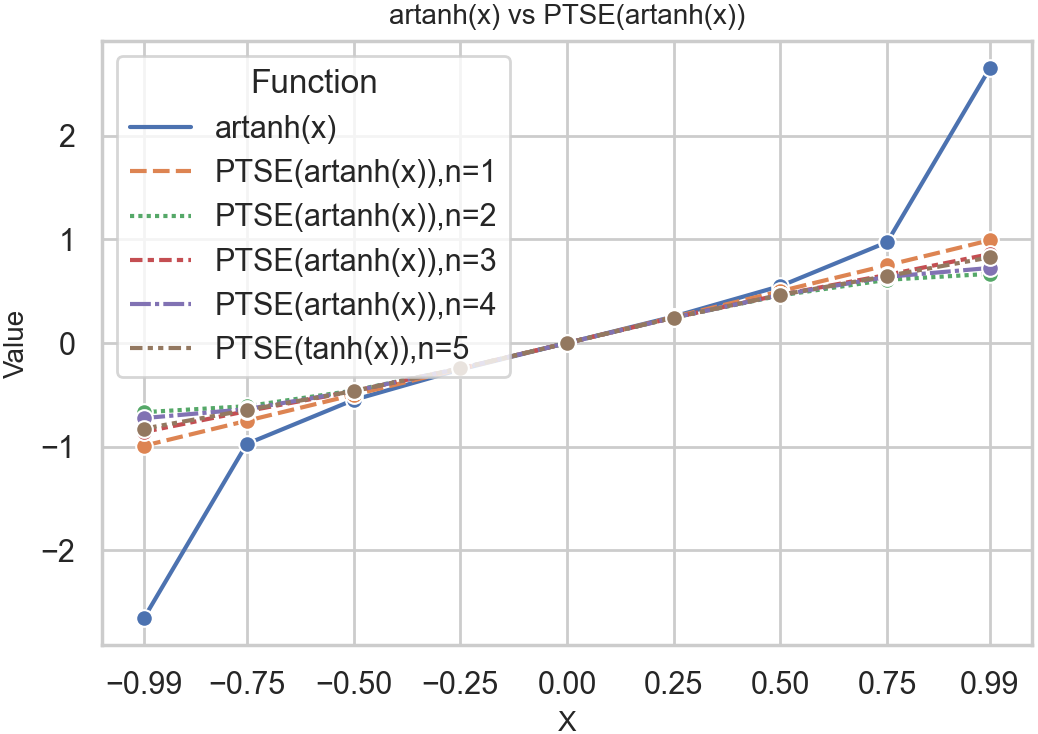}
        \caption{Plot of the $\artanh$ (solid line) function and its PTSE formulation at different values of precision.}
\end{subfigure}\hfill
\begin{subfigure}{.49\textwidth}
    \centering
    \includegraphics[width=\textwidth]{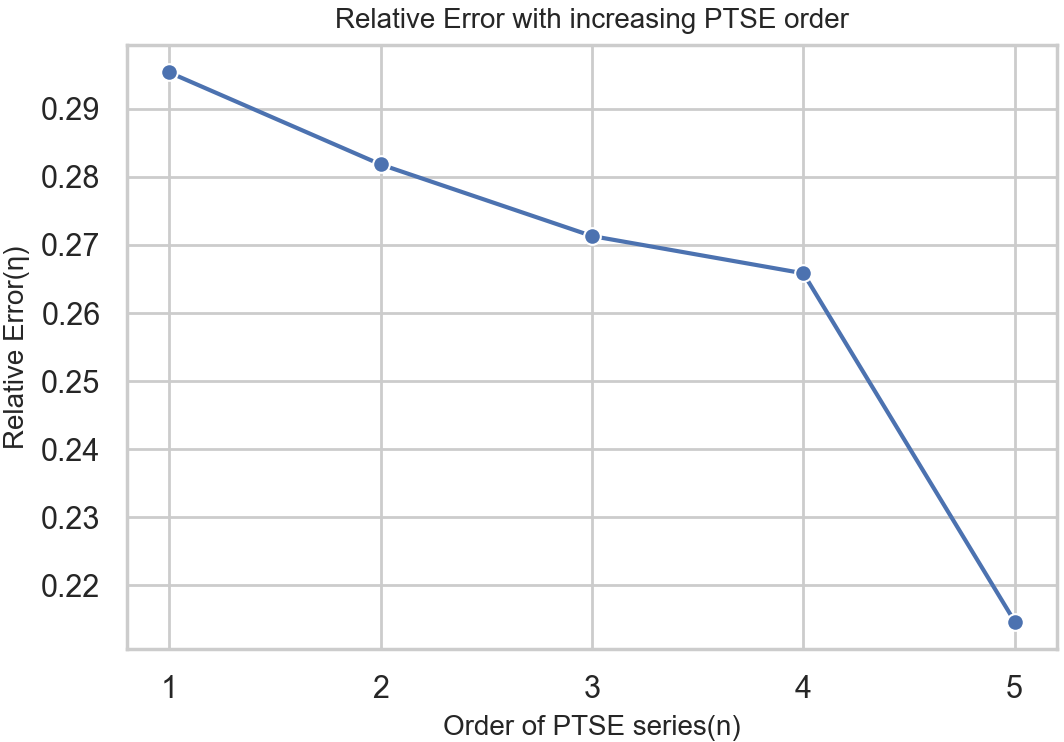}
        \caption{Relative Error between $\artanh$ and the PTSE formulation at different values of precision ($n$ vs $\eta$).}
\end{subfigure}
\begin{subfigure}{.49\textwidth}
    \centering
    \includegraphics[width=\textwidth]{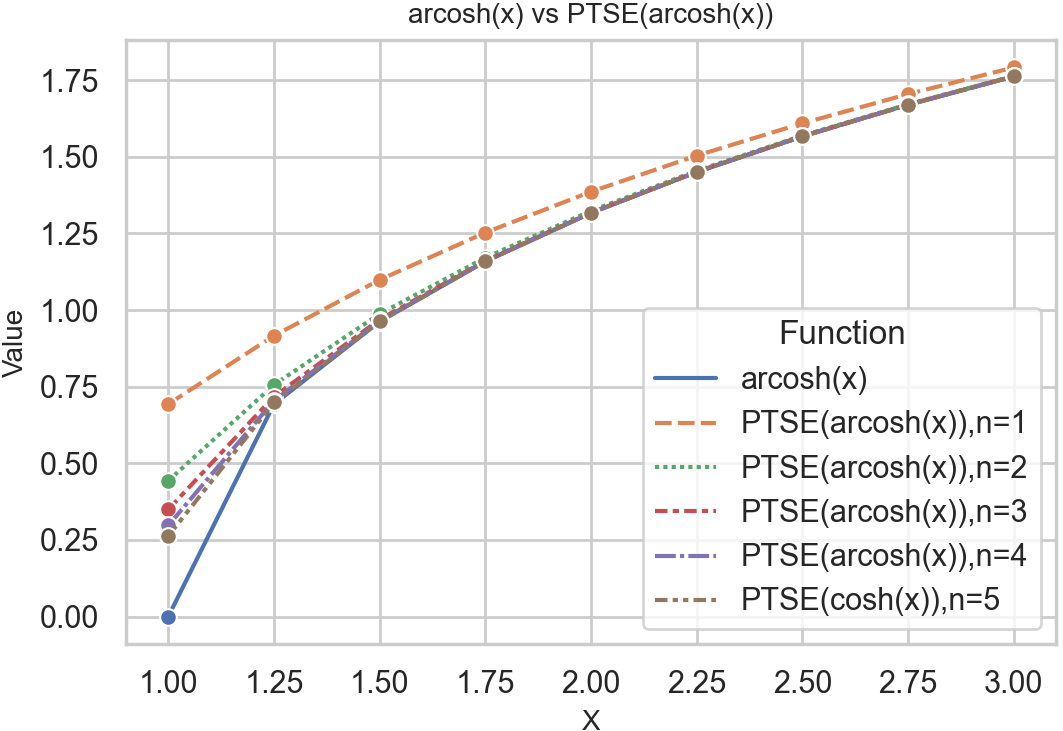}
        \caption{Plot of the $\arcosh$ (solid line) function and its PTSE formulation at different values of precision.}
\end{subfigure}\hfill
\begin{subfigure}{.466\textwidth}
    \centering
    \includegraphics[width=\textwidth]{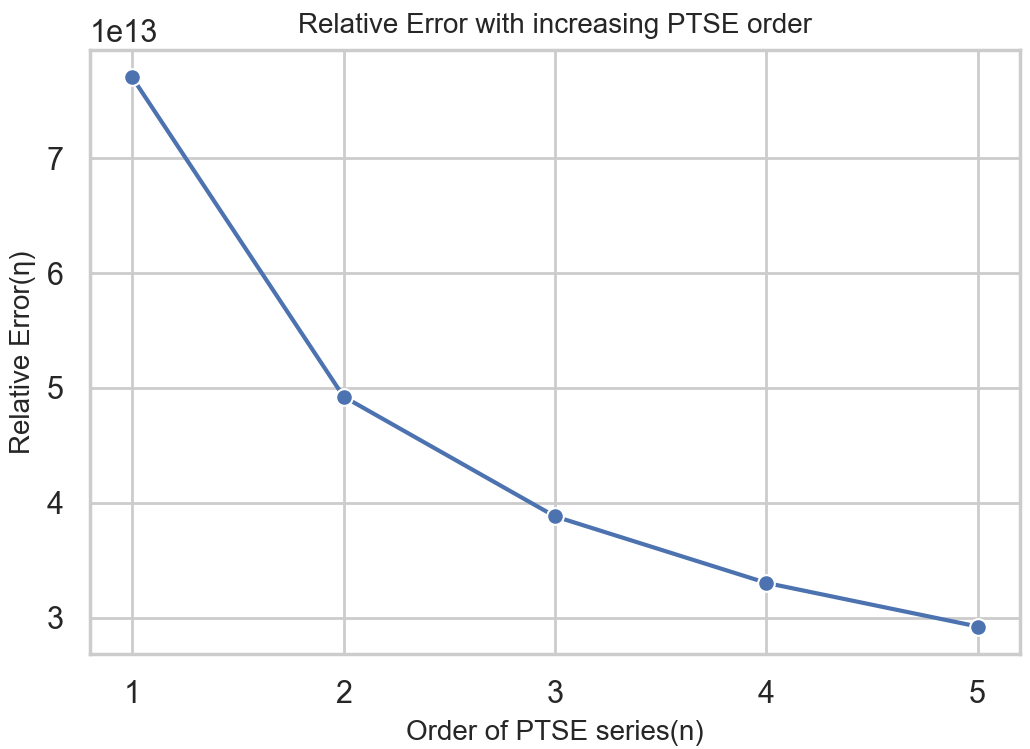}
        \caption{Relative Error between $\arcosh$ and the PTSE formulation at different values of precision ($n$ vs $\eta$).}
\end{subfigure}
\caption{Plots of hyperbolic trigonometric functions and their respective PTSE formulations and the relative error betweeen them at various precision values. We observe that the approximation error is the least for low values of the function $f(x)$ and high values of precision $n$.}
\label{fig:approximation}
\end{figure}
As illustrated in figure \ref{fig:approximation}, we note that approximation error is the least for low values of the function $f(x)$ and high values of precision $n$. However, the gains in precision also tend to saturate at a level of precision. Hence, we use a value-regularizer with a weight $\lambda$ to maintain low function values and tune the parameter of precision $n$ for efficient computation, faster convergence, and high performance.
\section{Case studies in Computer Vision and Natural Language Processing}
\label{app:case-study}
\begin{table}[htbp]
    \caption{Performance comparison results between MuRT (ours) and the baselines on the task of few shot image classification in the computer vision domain. The columns present the evaluation metric of Accuracy on the 1-shot 5 -way and 5-shot 5-way classification task. Best performing cells are highlighted in bold and second best are underlined.}
    \centering
    \begin{tabular}{ll|cc|cc}
    \hline
    \textbf{Dataset} & &\multicolumn{2}{c}{\textbf{MiniImageNet}} & \multicolumn{2}{c}{\textbf{Caltech-UCSD-Birds}}\\
    \textbf{Models} & \textbf{Backbone} & \textbf{1-shot 5-way} & \textbf{5-shot 5-way} & \textbf{1-shot 5-way} & \textbf{5-shot 5-way}\\\hline
    \textbf{ProtoNet} & \textbf{4Conv} & 49.42 & 68.20 &  51.31 & 70.77\\
         & \textbf{RSN12} & 56.50 & 74.2 & - & -\\\hline
    \textbf{H-ProtoNet} & \textbf{4Conv} & 54.43 & 72.67 & \textbf{64.02} & \textbf{82.53}\\
         & \textbf{RSN12} &  \textbf{59.47} & \textbf{76.84} & - & -\\\hline
     \textbf{T-ProtoNet} & \textbf{4Conv} & 53.54 & 70.18 & \underline{61.68}&\underline{77.46}\\
     & \textbf{RSN12} & \underline{58.02} & \underline{75.31} & - & -\\\hline
    \end{tabular}
    \label{tab:computer-vision}
\end{table}

\textbf{Computer Vision.} From the results provided in Table \ref{tab:computer-vision}, we observe that T-ProtoNet is able to perform significantly better than Euclidean ProtoNet and just slightly lower than the hyperbolic variant. This shows that T-ProtoNet is able to provide better results than Euclidean ProtoNet while being as scalable.

\begin{table}[htbp!]
\centering
    \caption{Performance comparison results between TSE formulation (ours) and the baselines on the task of textual entailment (SNLI) and noisy prefix detection (PREFIX-K\%) in the NLP domain. The columns present the evaluation metric of Accuracy for the different datasets. Best performing cells are highlighted in bold and second best are underlined.}
    \begin{tabular}{l|cccc}
    \hline
    \textbf{Models} & \textbf{SNLI} & \textbf{PREFIX-10\%} & \textbf{PREFIX-30\%} & \textbf{PREFIX-50\%}\\\hline
    \textbf{RNN} & 79.34 & 89.62 & 81.71 & 72.10\\
    \textbf{GRU} & \textbf{81.52} & 95.96 & 86.47 & 75.04\\\hline
    \textbf{HRNN} & 78.21 & 96.91 & 87.25 & 62.94 \\
    \textbf{HGRU} & 81.19 & \textbf{97.14} & \textbf{88.26} & \textbf{76.44}\\\hline
    \textbf{TRNN} & 80.19& 94.52& 82.53 & 72.87\\
    \textbf{TGRU} & \underline{81.41}&\underline{96.98} & \underline{87.40} & \underline{75.85}\\\hline
    \end{tabular}
\label{tab:nlp}
\end{table}
\textbf{Natural Language Processing.} From the results provided in Table \ref{tab:nlp}, we observe that PTSE formulations are able to perform significantly better than Euclidean counterparts and just slightly lower than the hyperbolic variants. This shows that PTSE is able to provide better results than Euclidean formulation while being as scalable.

\section{Simplification of PTSE HNN Components}
\label{app:simplification}
Let us assume a hyperbolic linear layer $f_l^\diamond(x)$ with weights $w$, inputs $x$ and bias $b$ is given by:
\begin{equation}
    f_l^\diamond(x) = w \otimes^\diamond x \oplus b
\end{equation}
Simplifying this using Eqs. (\ref{eq:prod}) and (\ref{eq:add}), we get
\begin{align}
    f_l^\diamond(x) &= K_{wx}wx \oplus b = K_{(wx,b)}(wx + b)
\end{align}
Here, $wx+b$ is a Euclidean linear layer $f_l(x)$ and $K_{(wx,b)}$ is a computationally efficient polynomial function. Hence, the hyperbolic linear layer with non-scalable hyperbolic trigonometric functions can be converted to Euclidean equivariants scaled by a polynomial function, using PTSE formulation, which are scalable on GPUs.

\section{Datasets and Baselines}
\label{app:datasets_baselines}
We select standard datasets, baselines, and evaluation metrics for each domain in the experiments; Graphs (GNNs), Multi-relational reasoning (MuR), Computer Vision (CV), and natural language processing (NLP). The datasets are summarized in Table \ref{tab:datasets} and described below:
\begin{itemize}[noitemsep,leftmargin=*]
    \item \textbf{Cora} (GNN) \cite{rossi2015the} contains 2708 publications classified into 7 classes and connected by citation links. 
    \item \textbf{Pubmed} (GNN) \cite{sen2008collective} is collected from the Pubmed database and contains 19717 papers classified into 3 classes and connected by citation links.
    \item \textbf{Citeseer} (GNN) \cite{sen2008collective} contains 3312 annotated scientific publications classified into 6 classes and connected by citation links. 
    \item \textbf{WN18RR} (MuR) \cite{dettmers2018convolutional} is a subset of the WordNet graph that connects words with semantic relations. WN18RR consists of 40,943 entities connected by 11 different semantic relations.
    \item \textbf{FB15k-237} (MuR) \cite{bordes2013translating, toutanova2015representing} contains triple relations from the Freebase entity pairs, with 14,541 entities connected by 237 relations.
    \item \textbf{MiniImageNet} (CV) \cite{russakovsky2015imagenet}: MiniImageNet dataset is the subset of ImageNet dataset with 600 examples per 100 of the sampled classes. From the experimental setup in \cite{khrulkov2020hyperbolic}, we set the training, validation, and test to consist of 64,16 and 20 classes, respectively. 
    \item \textbf{Caltech-UCSD-Birds} (CV) \cite{triantafillou2017few}: The CUB dataset consists of 11,788 images annotated with 200 bird classes. From the experimental setup in \cite{khrulkov2020hyperbolic}, we set the training, validation, and test to consist of 100, 50, and 50 classes, respectively.
    \item \textbf{SNLI} (NLP) \cite{bowman2015large}: The SNLI datasets consists of a pair of sentences, the premise and hypothesis, annotated by whether the hypothesis can be inferred from the premise or not. The dataset consists of 570K, 10K, and 10K training, validation, and test pairs, respectively. Following the experimental setup in \cite{ganea2018hyperbolic}, we use the classes of ``contradiction'' and ``neutral'' class as negative samples, and ``entailment'' as positive samples.
    \item \textbf{PREFIX-K\%} (NLP) \cite{ganea2018hyperbolic}: According to the methodology provided in \cite{ganea2018hyperbolic}, we use the K values of 10,30, and 50, where for each sentence, we replace K\% of the words of the prefix to get positive samples, and generate a random sentence of the same length as negative samples. With a word vocabulary size of 100, the training, validation, and test dataset contains  500K, 10K, and 10K samples, respectively.
\end{itemize}
The baseline methods (summarized in Table \ref{tab:datasets}) are described below:
\begin{itemize}[noitemsep,leftmargin=*]
    \item \textbf{GCN} (GNN) \cite{kipf2017semi}: Graph Convolution networks aggregate message from node's k-hop neighborhoods using learnable convolution filters.
    \item \textbf{GAT} (GNN)\cite{velickovic2018graph}: Graph Attention networks utilize attention weights to aggregate messages from the node neighborhood towards the target root node.
    \item \textbf{HGCN} (GNN) \cite{chami2019hyperbolic}: The Hyperbolic Graph Convolution network utilizes hyperbolic gyrovector formulation to aggregate the neighborhood's hyperbolic vectors at the root node's local tangent space for further message propagation. 
    \item \textbf{HGAT} (GNN) \cite{zhang2021hyperbolic}: The Hyperbolic Graph Attention networks utilizes the Lorentz model of hyperbolic space to aggregate node neighborhood in the hyperbolic space using Lorentz factors and Einstein Midpoint.
    \item \textbf{MuRE} (MuR) \cite{balazevic2019multi}: Multi-relational Euclidean model learns the representations of knowledge graph entitites and relations by formulating the tail entity as a translation of head entity by a relation vector. 
    \item \textbf{MuRP} (MuR) \cite{balazevic2019multi}: Similar to MuRE, the multi-relation Poincaré model learns the representation of KG entities and relations, but in the hyperbolic Poincaré ball by leveraging the gyrovector operations to formulate the hyperbolic translation of head entity by a relation vector.
    \item \textbf{ProtoNet} (CV) \cite{snell2017prototypical}: Focused on the problem of few-shot classification, Prototypical networks learn representations of each class in the training set and learn a metric space for classification on the test set that can be formed by computing distances from the training class representations.
    \item \textbf{H-ProtoNet} (CV) \cite{khrulkov2020hyperbolic}: H-ProtoNet is a hyperbolic variant of the ProtoNet model, formulated by essentially replacing the Euclidean mean operation with hyperbolic averaging (HypAve). 
    \item \textbf{RNN} (NLP) \cite{mikolov2010recurrent}: Recurrent Neural Networks are widely used models for sequential encoding that can learn dependencies between different timesteps of sequential inputs.
    \item \textbf{GRU} (NLP) \cite{junyoung2014empirical}: Gated recurrent networks are an optimization over RNNs to tackle the problem of exploding gradients using the reset gate and update gate.
    \item \textbf{HRNN} (NLP) \cite{ganea2018hyperbolic}: Hyperbolic RNNs utilize the hyperbolic linear layer to formulate the forward passes of RNNs and RAdam to perform the backpropagation over time for weight updates.
    \item \textbf{HGRU} (NLP) \cite{ganea2018hyperbolic}: Hyperbolic GRUs formulate the reset and update gate of the Euclidean GRUs in the hyperbolic space to learn hierarchical information.
\end{itemize}

\section{Evaluation Metrics}
The evaluation metrics for different problems (summarized in Table \ref{tab:datasets}) are described below:
\begin{itemize}[leftmargin=*]
    \item \textbf{Accuracy in node classification:} The metric computes the ratio of the number of correct predictions of node classes and the total number of nodes in the test set, i.e., $Accuracy = \frac{\# \text{number of correctly classified nodes}}{\# \text{total nodes}}$
    \item \textbf{Area under ROC in link prediction:} The metric is a quantification of the ability of classifiers to distinguish between classes on the basis of signal and noise. For a binary classification problem, this can be computed as; $AU-ROC = \frac{1+TPR-FPR}{2}$, where $TPR$ and $FPR$ are the true positive and false positive rates, respectively. 
    \item \textbf{HITS@K in multi-relation reasoning:} The metric is used to compute the precision of correct results in the the top K results of a query with head entity and relation, i.e., $HITS@K = \frac{1}{K}\sum_{i=1}{K}f(e_i)$, where $f(e_i)=1$, if $e_i$ is a correct result, and $f(e_i)=0$ otherwise.
    \item \textbf{Mean reciprocal rank in multi-relation reasoning:} The metric is used to compute the rank of correct results retrieved set of answers, i.e., $MRR = \frac{1}{n}\sum_{i=1}{n}\frac{1}{f(e_i)}$, where $n$ is the total number of samples, and $f(e_i)=1$, if $e_i$ is a correct result, and $f(e_i)=0$ otherwise.
    \item \textbf{Accuracy in image classification:} The metric computes the ratio of the number of correct predictions of image classes and the total number of samples in the test set, i.e., $Accuracy = \frac{\# \text{number of correctly classified images}}{\# \text{total size of test set}}$
    \item \textbf{Accuracy in textual entailment and noisy prefix detection:} The metric computes the ratio of the number of correctly classified sentence pairs of (premise, hypothesis), for textual entailment, and (noisy prefix, sentence), for noisy prefix detection, and the total number of samples in the test set, i.e., $Accuracy = \frac{\# \text{number of correctly classified sentence pairs}}{\# \text{total size of test set}}$
    
\end{itemize}